\documentclass[preprint,12pt]{elsarticle}

\usepackage{amssymb}
\usepackage{amsmath}
\usepackage{algorithmic}
\usepackage{graphicx}
\usepackage{textcomp}
\usepackage{booktabs}
\usepackage{makecell}
\usepackage{multirow}
\usepackage{url}
\usepackage{longtable}
\usepackage{colortbl}
\usepackage{wrapfig}
\usepackage{float}
\usepackage{tikz}
\usepackage{forest}

\usepackage{xcolor}
\definecolor{mygray}{gray}{.9}
\definecolor{mypink}{rgb}{.99,.91,.95}
\definecolor{mycyan}{cmyk}{.6,0,0,0}
\definecolor{mygreen}{rgb}{0,1,0}

\usepackage{wasysym}

\newcommand{\etal}{\textit{et al.} }
\journal{Nuclear Physics B}

\begin{document}

\begin{frontmatter}



\title{Self-eXplainable AI for Medical Image Analysis:\\ A Survey and New Outlooks}


\author[label1]{Junlin Hou} %% Author name
\affiliation[label1]{%
  organization={Department of Computer Science and Engineering, Hong Kong University of Science and Technology},
  city={Hong Kong},
  country={China}}

\author[label2]{Sicen Liu} %% Author name
\affiliation[label2]{%
  organization={Shenzhen MSU-BIT University},
  city={Shenzhen},
  country={China}}

\author[label1]{Yequan Bie}
\author[label1]{Hongmei Wang}
\author[label1]{Andong Tan}

\author[label1]{Luyang Luo} %% Author name
% \affiliation[label3]{%
%   organization={Harvard University},
%   city={Cambridge},
%   country={USA}}

\author[label1,label4,label5,label6,label7]{Hao Chen \corref{cor1}}
\affiliation[label4]{%
  organization={Department of Chemical and Biological Engineering, Hong Kong University of Science and Technology},
  city={Hong Kong},
  country={China}}
\affiliation[label5]{%
  organization={Division of Life Science, Hong Kong University of Science and Technology},
  city={Hong Kong},
  country={China}}
\affiliation[label6]{%
  organization={HKUST Shenzhen-Hong Kong Collaborative Innovation Research Institute, Futian},
  city={Shenzhen},
  country={China}}
\affiliation[label7]{%
  organization={State Key Laboratory of Nervous System Disorders, The Hong Kong University of Science and Technology},
  city={Hong Kong},
  country={China}}
\cortext[cor1]{Corresponding author}

% \affiliation{organization={},%Department and Organization

%% Abstract
\begin{abstract}
The increasing demand for transparent and reliable models, particularly in high-stakes decision-making areas such as medical image analysis, has led to the emergence of eXplainable Artificial Intelligence (XAI). Post-hoc XAI techniques, which aim to explain black-box models after training, have raised concerns about their fidelity to model predictions. In contrast, Self-eXplainable AI (S-XAI) offers a compelling alternative by incorporating explainability directly into the training process of deep learning models. This approach allows models to generate inherent explanations that are closely aligned with their internal decision-making processes, enhancing transparency and supporting the trustworthiness, robustness, and accountability of AI systems in real-world medical applications. To facilitate the development of S-XAI methods for medical image analysis, this survey presents a comprehensive review across various image modalities and clinical applications. It covers more than 200 papers from three novel perspectives: 1) input explainability through the integration of explainable feature engineering and knowledge graph, 2) model explainability via attention-based learning, concept-based learning, and prototype-based learning, and 3) output explainability by providing textual explanations and visual counterfactual explanations. This paper also outlines desirable characteristics of S-XAI models for assessing explanation quality, while discussing major challenges and future research directions in developing S-XAI for medical image analysis.
\end{abstract}

% \includegraphics[width=\linewidth]{LaTeX/fig/junlin/graphabs.pdf}

% \begin{highlights}
% \item First comprehensive survey on Self-eXplainable AI (S-XAI) in medical image analysis.

% \item Novel three-level taxonomy: input, model, and output explainability.

% \item In-depth analysis of 200+ publications (2018–2025) across top venues.

% \item Comprehensive review of methods, medical applications, and evaluations.

% \item Identifies key challenges and outlines actionable future research directions.

% \end{highlights}

%% Keywords

\end{frontmatter}



\section{Introduction}
Artificial intelligence (AI), particularly deep learning, has driven significant advancements in medical image analysis, including applications in disease diagnosis, lesion segmentation, medical report generation (MRG), and visual question answering (VQA). Deep neural networks (DNNs) automatically learn features from input data and produce optimal outputs. However, the inherently complex nature of DNNs hinders our understanding of the decision-making processes behind these models. Consequently, DNNs are often considered as black-box models, which has raised concerns about their transparency, interpretability, and accountability for successful deployments in real-world clinical scenarios \cite{jia2020clinical}. 

\begin{figure}[t]
\begin{center}
   \includegraphics[width=\linewidth]{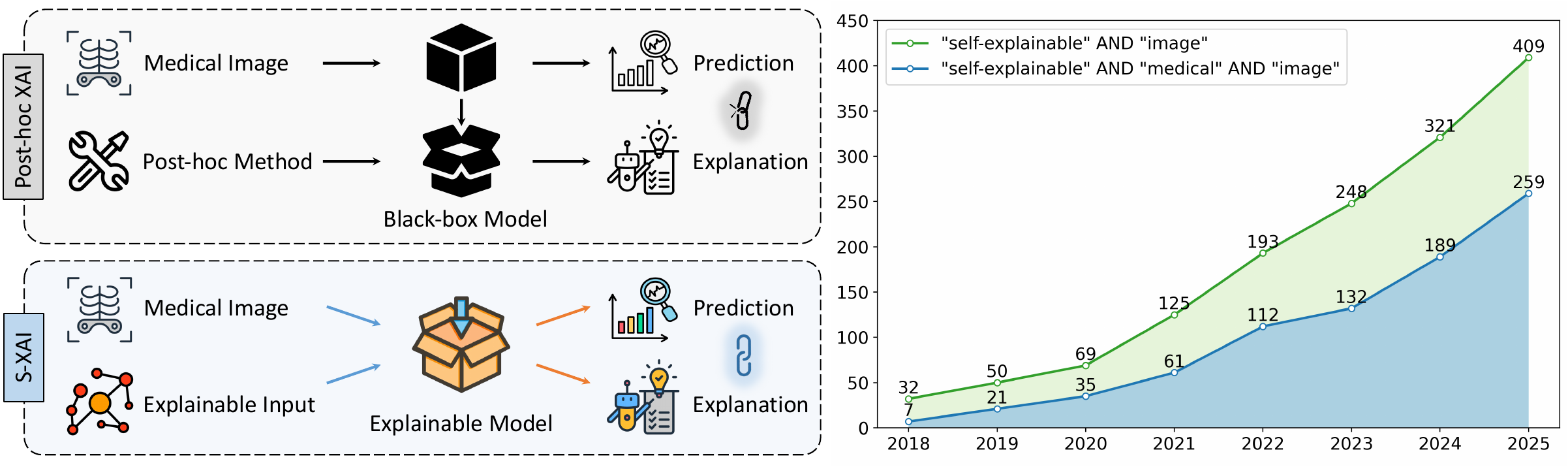}
\end{center}
% \vspace{-1.0em}
   \caption{Left: Illustration of post-hoc XAI versus Self-eXplainable AI (S-XAI). Right: The upward trend of the total number of S-XAI research papers from 2018 to 2025, where medical papers account for half.}
\label{fig:bar}
% \vspace{-1.0em}
\end{figure}

To tackle the challenge of developing more trustworthy AI systems, research efforts are increasingly focusing on various eXplainable AI (XAI) methods, enhancing transparency, fairness, and robustness.
However, most XAI methods aim to generate explanations for the outputs of black-box AI models after they have been trained, a category known as post-hoc XAI, as illustrated in Fig. \ref{fig:bar} left top. 
These methods utilize additional explanation models or algorithms to provide insights into the decision-making process of the primary AI model.
In the field of medical image analysis, commonly used post-hoc XAI techniques include feature attribution methods, such as gradient-based approaches (e.g., LRP \cite{bach2015pixel}, CAM \cite{zhou2016learning}) and perturbation-based approaches (e.g., LIME \cite{ribeiro2016should}, Kernel SHAP \cite{lundberg2017unified}). Additionally, some methods explored concept attributions, learning human-defined concepts from the internal activations of DNNs (e.g., TCAV \cite{tcav}, CAR \cite{crabbe2022concept}).
Post-hoc XAI techniques are often model-agnostic, indicating that they can be flexibly applied to a variety of already-trained black-box AI models.

Since post-hoc explanations are generated separately from the primary AI model, several valid concerns have been raised: 1) These explanations may not always be faithful to the actual decision-making process of black-box models \cite{mittelstadt2019explaining,zhang2022overlooked}; 2) They may lack sufficient detail to fully elucidate the model's functioning  \cite{rudin2019stop}. 
These limitations of post-hoc XAI approaches are particularly problematic in high-stakes domains like medical image analysis, where clinicians require a deep and trustworthy understanding of how an AI model arrives at its predictions. The issues about the faithfulness and sufficiency of post-hoc explanations highlight the importance of exploring self-explainable AI models as a potentially more reliable and transparent alternative.

Self-eXplainable AI (S-XAI) is a category of XAI methods designed to be interpretable by nature, as illustrated in Fig. \ref{fig:bar} left bottom. These methods incorporate explainability as an integral part of the model during the training process, rather than generating explanations after the model has been trained.
Conventional inherently interpretable methods include various white-box machine learning models, such as decision trees \cite{quinlan1986induction}, generalized additive models \cite{hastie2017generalized}, and rule-based systems \cite{grosan2011intelligent}. 
In this survey, we focus primarily on DNNs and analyze the characteristics of self-explainability across the entire pipeline, from model input to architecture to output, enabling direct inspection and understanding of the reasoning behind the model's predictions without reliance on external explanation methods.
In contrast to post-hoc XAI approaches, S-XAI methods aim to provide explanations that are inherent, transparent, and faithful, aligning directly with the model's internal decision-making mechanisms. Such explanations are essential for the effective adoption and clinical integration of AI-powered decision support systems. Furthermore, S-XAI facilitates collaborative decision-making between clinicians and AI systems, fostering better-informed and more accountable medical diagnoses and interventions. 

This paper presents the first systematic review of S-XAI for medical image analysis, covering methodology, medical applications, and evaluations, while also offering an in-depth discussion on challenges and future directions. 
Although there is a wealth of literature on medical XAI surveys \cite{patricio2023explainable,salahuddin2022transparency,yang2022unbox,van2022explainable,sheu2022survey,tjoa2020survey,singh2020explainable,hossain2025explainable} that deliver valuable insights, 
none have focused specifically on a comprehensive review of S-XAI methods applied to medical image analysis.
Table \ref{tab:survey_comparison} positions this survey among the existing medical XAI surveys in terms of the three explainability layers.

\begin{table*}[!t]
\centering
\caption{Comparison of the present survey with existing medical XAI surveys. FE: feature engineering; KG: knowledge graph; CF: counterfactual. $\checkmark$: the survey covers the corresponding explainability methods; $\times$: not (systematically) covered.}
\label{tab:survey_comparison}
\scriptsize
\setlength{\tabcolsep}{2pt}
\resizebox{\textwidth}{!}{%
\begin{tabular}{@{}l l cc ccc cc@{}}
\toprule
\multirow{2}{*}{Survey} & \multirow{2}{*}{Year} & \multicolumn{2}{c}{Input Explainability} & \multicolumn{3}{c}{Model Explainability} & \multicolumn{2}{c}{Output Explainability} \\
\cmidrule(lr){3-4}\cmidrule(lr){5-7}\cmidrule(lr){8-9}
 & & FE & KG & Attention & Concept & Prototype & Textual & Visual CF \\
\midrule
Tjoa \& Guan \cite{tjoa2020survey} & 2020 & \checkmark & $\times$ & $\times$ & $\times$ & $\times$ & \checkmark & $\times$ \\
Singh \etal \cite{singh2020explainable} & 2020 & $\times$ & $\times$ & \checkmark & $\times$ & $\times$ & \checkmark & $\times$ \\
Salahuddin \etal \cite{salahuddin2022transparency} & 2022 & $\times$ & $\times$ & \checkmark & \checkmark & \checkmark & \checkmark & $\times$ \\
Van der Velden \etal \cite{van2022explainable} & 2022 & $\times$ & $\times$ & \checkmark & $\times$ & \checkmark & \checkmark & $\times$ \\
Sheu \& Pardeshi \cite{sheu2022survey} & 2022 & $\times$ & $\times$ & $\times$ & $\times$ & $\times$ & $\times$ & $\times$ \\
Yang \etal \cite{yang2022unbox} & 2022 & \checkmark & $\times$ & \checkmark & $\times$ & $\times$ & $\times$ & $\times$ \\
Patr\'icio \etal \cite{patricio2023explainable} & 2023 & $\times$ & $\times$ & $\times$ & \checkmark & \checkmark & \checkmark & \checkmark \\
Hossain \etal \cite{hossain2025explainable} & 2025 & $\times$ & $\times$ & \checkmark & $\times$ & \checkmark & \checkmark & \checkmark \\
\textbf{Ours} & 2026 & \checkmark & \checkmark & \checkmark & \checkmark & \checkmark & \checkmark & \checkmark \\
\bottomrule
\end{tabular}%
}
\renewcommand{\tabcolsep}{6pt}
\end{table*}

We analyze more than 200 papers published from 2018 to 2025, sourced from the proceedings of NeurIPS, ICLR, ICML, AAAI, CVPR, ICCV, ECCV, and MICCAI as well as top-tier journals in the field, including Nature, Nature Machine Intelligence, Nature Medicine, Nature Communications, Information Fusion, IEEE TPAMI, IEEE TMI, MedIA, or those cited in related works.
For literature retrieval, we performed keyword queries on Google Scholar using terms such as \textit{self-explainable}, \textit{feature engineering}, \textit{knowledge graph}, \textit{attention-based}, \textit{prototype-based}, \textit{concept-based}, \textit{textual explanation}, \textit{counterfactual explanation}, and \textit{medical image analysis}, which covers the venues and journals listed above. Candidates were screened by title and abstract first, and then reviewed in full text. A paper was included if it embeds explainability intrinsically into the model and addresses medical image analysis. The reference lists of the included papers were further screened to capture additional relevant works.
The statistics of research articles using keywords \textit{self-explainable}, \textit{medical}, \textit{image} on Google Scholar are presented in Fig. \ref{fig:bar} right, which reveal two key observations:
1) There has been a significant and consistent increase in research papers focused on self-explainable AI over the years, indicating growing interest and emphasis within the research community;
2) Nearly half of the total research papers (green areas) are dedicated to applying S-XAI techniques in medical image analysis (blue areas), highlighting the vital importance of S-XAI in the medical field.

To summarize, this survey presents insights into S-XAI for medical image analysis, with five main contributions outlined below:
\begin{enumerate}
    \item \textbf{Novel scope of XAI survey for medical image analysis:} As an emerging XAI paradigm, S-XAI integrates explainability intrinsically into the model itself, attracting growing research attention. This work represents the first comprehensive survey on this topic.
    \item \textbf{Systematic review of methods:} We introduce a novel taxonomy of relevant papers and review them based on input explainability, model explainability, and output explainability. This offers insights into potential technical innovations for S-XAI methods.
    \item \textbf{Thorough overview of applications:} We overview various downstream applications across different anatomical locations and modalities in current S-XAI research. This illustrates the ongoing development of S-XAI technologies in medical image analysis, serving as a reference for future applications in diverse contexts.
    \item \textbf{Comprehensive survey of evaluations:} We analyze a range of desired characteristics and evaluation methods to assess the quality of explainability. This provides guidelines for developing clinically explainable AI systems that are trustworthy and meaningful for end-users.
    \item \textbf{In-depth discussion of challenges and future work:} We discuss the key challenges and look forward to the promising future directions. 
    This underscores existing shortcomings and identifies new outlooks for researchers to drive further advancements.
\end{enumerate}

\section{S-XAI in Medical Image Analysis}

Transparency and trustworthiness are essential for deep learning models deployed in real-world applications of medical image analysis. To address this need, the research community has explored various XAI methods and proposed several XAI taxonomies. 
According to existing literature \cite{patricio2023explainable,van2022explainable,christoph2020interpretable}, XAI methods can be categorized by the following criteria.

\begin{enumerate}
    \item \textbf{Intrinsic versus Post-hoc:} This criterion differentiates whether interpretability is inherent to the model's architecture (intrinsic) or achieved after the model training (post-hoc). 
    \item \textbf{Model-specific versus Model-agnostic:} Model-specific methods are restricted to particular model classes, whereas model-agnostic methods can be applied to explain any model.
    \item \textbf{Local versus Global:} The scope of an explanation distinguishes between those for an individual prediction (local) or those for the entire model behavior (global).
    \item \textbf{Explanation modality:} The common types of explanation include visual explanation, textual explanation, concept explanation, sample explanation, etc.
\end{enumerate}

\begin{figure}[H]
\begin{center}
   \includegraphics[width=\linewidth]{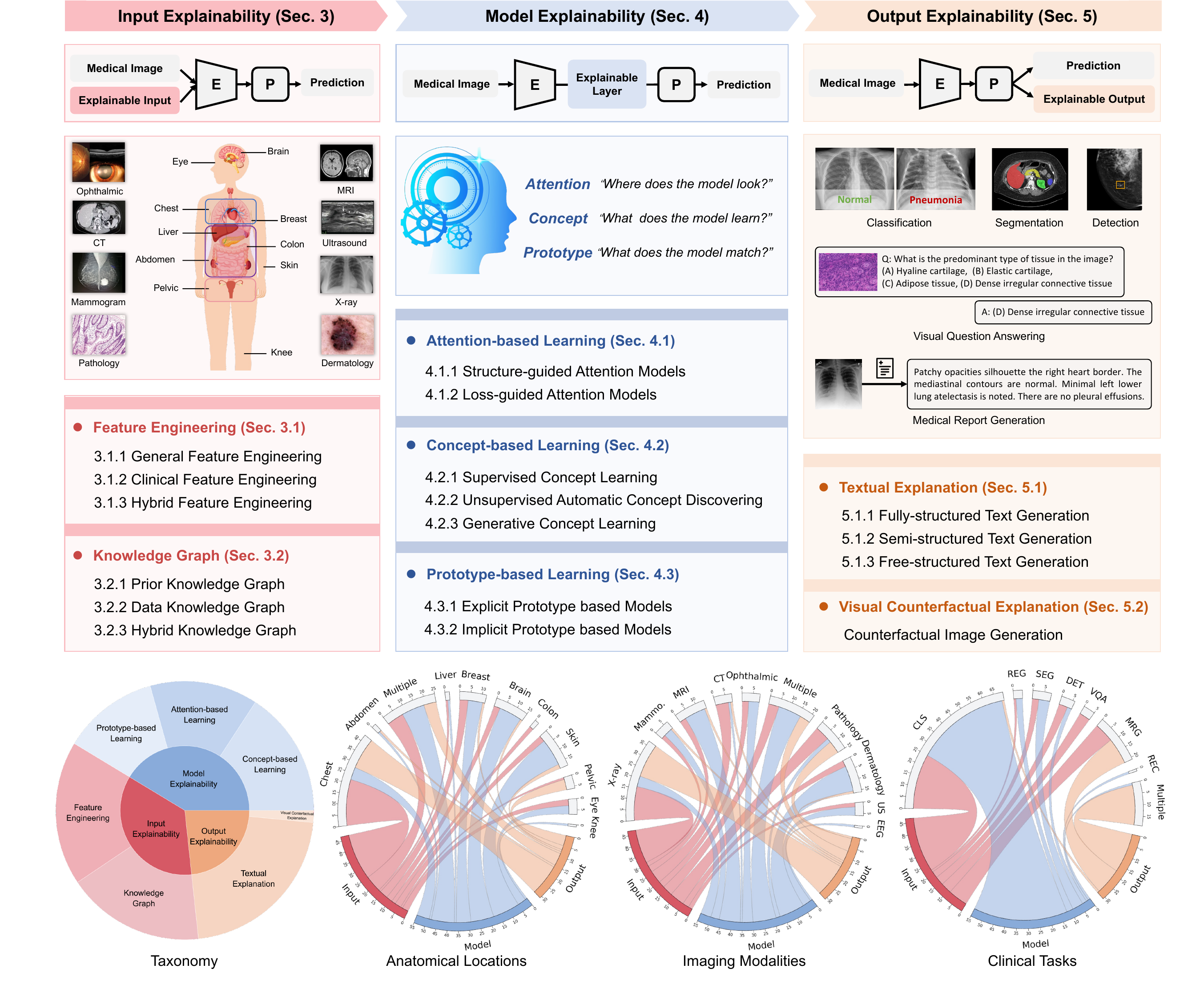}
\end{center}
% \vspace{-1.0em}
   \caption{Overview of Self-eXplainable AI (S-XAI) frameworks for DNNs, including input explainability, model explainability, and output explainability.}
\label{fig:methodology}
% \vspace{-1.0em}
\end{figure}

This survey concentrates on S-XAI methods for medical image analysis that allow models to inherently explain their own decision-making. 
As illustrated in Fig. \ref{fig:methodology}, we introduce a new taxonomy of S-XAI methods structured around the three fundamental components of DNNs. Additionally, we present statistics on the distribution of reviewed papers across the proposed taxonomy, along with chord diagrams that visualize the relationships between the three types of S-XAI approaches and various anatomical locations, imaging modalities, and clinical tasks.

\begin{enumerate}
    \item \textbf{Input explainability (Sec. \ref{sec:input}):} Input explainability focuses on integrating additional explainable inputs with deep features of medical images obtained from various anatomical locations and modalities to produce final predictions. By incorporating external knowledge and context-specific information, the accuracy and reliability of these predictions can be significantly improved.
    \item \textbf{Model explainability (Sec. \ref{sec:model}):} Model explainability aims to  design inherently interpretable model architectures of DNNs. Instead of explaining a black-box model, transforming the model into an interpretable format enhances understanding of how it processes information.
    \item \textbf{Output explainability (Sec. \ref{sec:output}):} Output explainability refers to the model's ability to generate not just predictions for various medical image tasks but also accompanying explanations through an explanation generator. This capability aids in understanding the rationale behind the model's predictions, facilitating informed medical decision-making.
\end{enumerate}

\noindent It is worth noting that these three categories are defined according to the primary locus where explainability is introduced in the deep learning pipeline. In practice, some methods may naturally span multiple dimensions. For example, concept-based models may ultimately express their learned concepts through textual explanations, while certain S-XAI systems combine interpretable inputs, structured internal reasoning, and explanatory outputs within a unified framework. Throughout this survey, we categorize each method according to its dominant explainability mechanism while explicitly acknowledging possible overlaps where relevant.

The following sections summarize and categorize the most relevant works on S-XAI methods applied to medical image analysis.
Comprehensive lists of the reviewed S-XAI methods are provided, detailing the employed S-XAI techniques, publication year, anatomical location, image modality, medical application, and the datasets used. Detailed information about the datasets is available in the supplementary material.

\section{Input Explainability} \label{sec:input}

\begin{figure}[t]
\begin{center}
   \includegraphics[width=\linewidth]{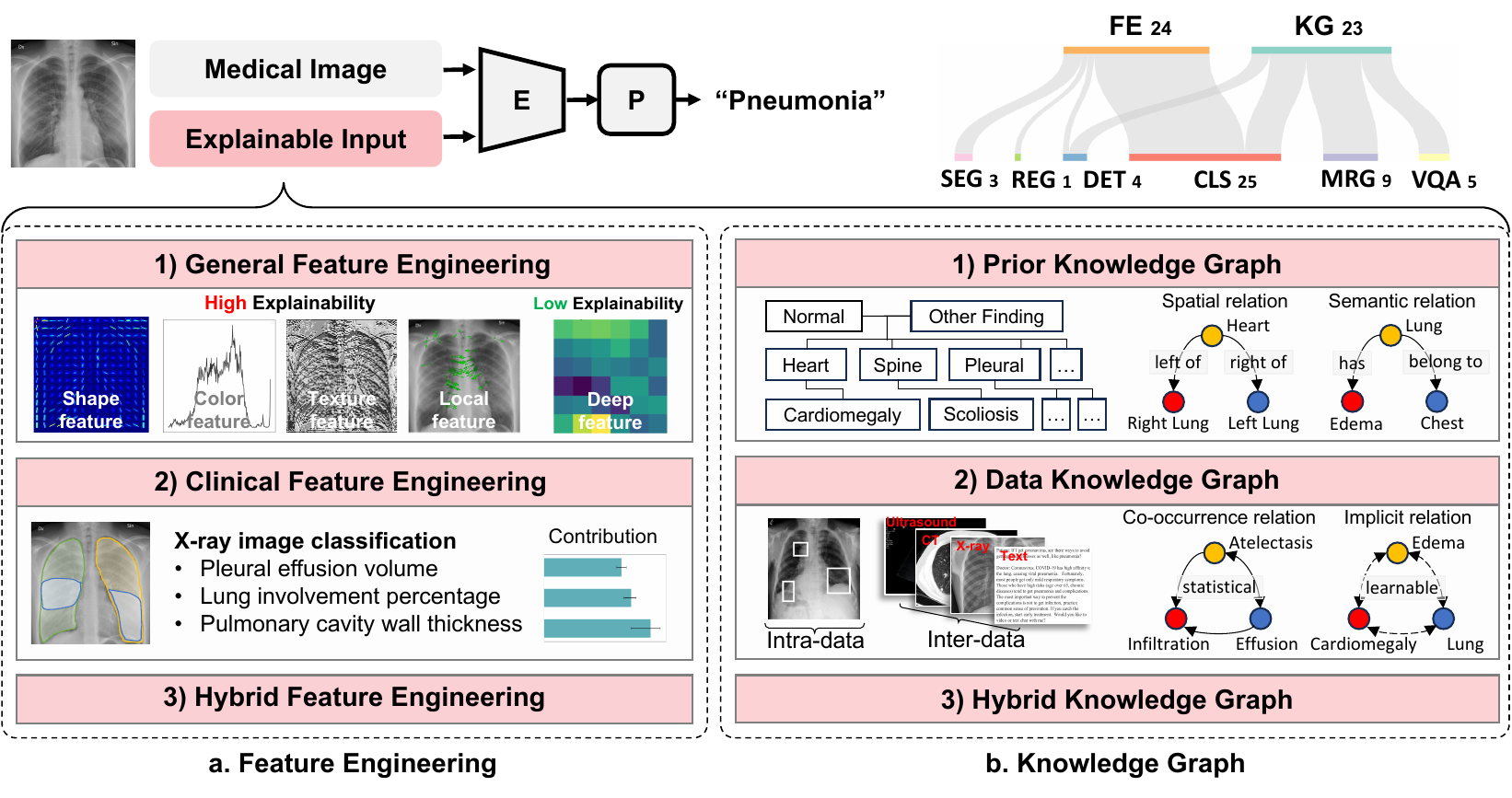}
\end{center}
% \vspace{-1.0em}
   \caption{Input explainability that incorporates explainable feature engineering and knowledge graphs as additional inputs.}
\label{fig:input_exp}
% \vspace{-1.0em}
\end{figure}

Input explainability is defined as an approach that enhances model transparency by integrating additional, interpretable inputs alongside raw medical images. In this section, we explore two methods to achieve this, i.e, feature engineering (Sec. \ref{subsec:input-1}) and knowledge graph (Sec. \ref{subsec:input-2}). As illustrated in Fig. \ref{fig:input_exp}, these explainable inputs are integrated with deep visual features to jointly inform the final prediction, enhancing both the accuracy and interpretability of the decision-making process. 
The Sankey diagram (upper right) highlights that feature engineering is primarily employed for recognition-oriented tasks, such as classification, segmentation, and detection, whereas knowledge graphs are more frequently applied to reasoning-intensive tasks, including medical report generation and visual question answering.

\subsection{Feature Engineering} \label{subsec:input-1}

Deep learning automates feature extraction from raw images, but these features often lack clinical interpretability. In contrast, feature engineering aims to transform raw images into a set of human-interpretable features, despite being more time-consuming \cite{zheng2018feature}. To enhance the explainability of deep learning models, it is beneficial to integrate explainable feature engineering with deep learning for decision making. 
As illustrated in Fig. \ref{fig:input_exp}a, we categorize feature engineering into three types: 1) General feature engineering, which employs universal image-processing algorithms to quantify fundamental visual attributes; 2) Clinical feature engineering, which quantifies pathology-aware attributes to encode high-level semantic information directly relevant for clinical decision-making; and 3) Hybrid feature engineering, which combines the advantages of both general and clinical approaches. A comprehensive list of input explainability methods based on these feature engineering approaches for medical image analysis is provided in Table \ref{tab:feature}.

\subsubsection{General feature engineering}

Compared to the deep features learned by neural networks, general feature engineering yields features that are more transparent and easily interpretable to humans. It involves quantifying fundamental visual attributes using predefined, domain-agnostic algorithms to capture core image aspects. For example, shape can be characterized by Histogram of Oriented Gradients (HOG), color by color moments and histograms, texture by Gray-Level Co-occurrence Matrix (GLCM) or Local Binary Patterns (LBP), and local keypoints by detectors such as SIFT or SURF.

In the field of medical image analysis, strategies that fuse deep learning features with these diverse handcrafted features have been widely applied to classification tasks involving images of dermatology \cite{majtner2016combining,sharma2022dermatologist}, pathology \cite{kumar2022empirical}, MRI \cite{khan2020cascading,saba2020brain}, CT \cite{li2019predicting,hu2024ct}, X-ray \cite{prince2025interpretable}, and others \cite{zhang2017classification,li2019integrating,daoud2020breast,sajid2023breast}.
For instance, Hu \etal \cite{hu2024ct} combine 1,967 handcrafted radiomic features with deep features from VGG16 to predict the response to anti-PD-1 therapy in NSCLC from CT images. 
Among the nine most important predictors identified, seven were handcrafted, five of which fell into the textural feature category. This highlights the dominant predictive value of engineered features and suggests that tumor complexity and heterogeneity may be critically associated with treatment outcome.
Similar fusion approaches have also been explored in medical image segmentation tasks \cite{manivannan2017structure,khan2020cascading,mecheter2022deep} and detection tasks \cite{kashif2016handcrafted}, where general handcrafted features can provide complementary contour and texture information to refine object boundaries and localization.
However, general feature engineering captures low-level patterns but lacks clinical semantics, thus prompting clinical-related approaches to bridge this gap with medical domain knowledge.

\subsubsection{Clinical feature engineering}

Clinical feature engineering provides a pathway to model explanations that are inherently aligned with clinical reasoning. It achieves this by quantifying disease-specific semantic attributes from medical knowledge, thereby formalizing diagnostic observations into computational features. The contribution of these features to a model's decision can then be transparently quantified using feature importance methods.

In CT imaging, Lou \etal \cite{lou2019image} and Chen \etal \cite{chen2025interpretable} utilize PyRadiomics to extract handcrafted radiomic features, which were then integrated with deep learning features for treatment response prediction and survival prediction, respectively. To enhance model transparency, Chen \etal \cite{chen2025interpretable} further employ SHapley Additive exPlanations (SHAP) to identify the key factors influencing the predictions. Similarly, in pathology images, Diao \etal \cite{diao2021human} predict molecular phenotypes using 607 human-interpretable features (HIFs) derived from cell and tissue segmentation. These HIFs, encompassing metrics from simple cell counts to complex spatial relationships, established a transparent link to clinically relevant phenotypes, thereby facilitating biological insight. SI-MIL \cite{kapse2024si} also contributes to this area by introducing a self-interpretable framework that jointly learns from both deep features and handcrafted morphometric and spatial descriptors, demonstrating interpretability at both local and global levels.

The fusion of medical images with clinical and biological data represents a powerful extension of this paradigm, enhancing model's performance and clinical relevance. 
This is evidenced by Lassau \etal \cite{lassau2021integrating}, who show that a model predicting COVID-19 severity learned to internalize clinical features from CT scans. Similarly, for ovarian cancer diagnosis, OvcaFinder  \cite{xiang2024development} uses SHAP values to quantify each input feature's contribution, ensuring transparency in individual predictions.
Nevertheless, implementing clinical feature engineering faces challenges such as its reliance on extensive domain expertise, potential subjectivity, and limited generalizability, underscoring the need for a hybrid approach that leverages both general and clinical-specific paradigms.

\subsubsection{Hybrid feature engineering}

Hybrid feature engineering represents an integrative approach that combines general and clinical feature engineering, often fusing them with deep learning representations. This strategy not only enhances predictive performance but also contributes to improved model interpretability by embedding a more comprehensive set of human-understandable concepts alongside powerful deep features.

For instance, in fundus image analysis, Harangi \etal \cite{harangi2019automatic} fuse intensity, geometry, texture, and lesion-specific features (such as the number of microaneurysms and exudate-specific attributes) with deep features from AlexNet for improved ocular disease classification. Similarly, in dermatology, Hagerty \etal \cite{hagerty2019deep} integrate color and texture features, salient points, telangiectasia vessel attributes, atypical pigment network patterns, and demographic data to advance melanoma classification.
Cruz \etal \cite{cruz2023benign} develop a hybrid feature set for breast tumor classification, which combined deep learning features, handcrafted descriptors (HOG, LBP), and perceptual features derived from Breast Imaging Reporting and Data System (BI-RADS).

\textbf{Discussion:} 
From the perspective of input explainability, feature engineering serves as a crucial bridge between data-driven models and human domain knowledge, ensuring that models base their reasoning on medically meaningful evidence starting from the input level.  Future research should therefore move beyond pure performance metrics and prioritize the creation of interpretable, verifiable input representations. Such evolution promises to transform deep learning from opaque black boxes into clinically validated decision-support systems founded on trustworthy features.

% 有数据集名称，无数据集引用
\begin{table*}[t]
\centering
\caption{Input explainability methods based on feature engineering (FE). The abbreviations here are CLS: classification, DET: detection, SEG: segmentation, REG: regression.}
\resizebox{\columnwidth}{!}{
\begin{tabular}{@{}lcccccc@{}}
% \begin{tabular}{ccccc}
\toprule
\textbf{Method} & \textbf{Year} & \textbf{Location} & \textbf{Modality} & \textbf{Task} & \textbf{Dataset} & \textbf{FE Type} \\
\midrule

Majtner \etal \cite{majtner2016combining} & 2016 & Skin & Dermatology & CLS & ISIC 2016 & General FE\\

Sharma \etal \cite{sharma2022dermatologist} & 2022 & Skin & Dermatology & CLS & HAM10000 & General FE \\

% Kumar \etal \cite{kumar2022empirical} & 2022 & Lung, Colon & Pathology & CLS & \cite{borkowski2019lung} & General FE \\ %xai
Kumar \etal \cite{kumar2022empirical} & 2022 & Multiple & Pathology & CLS & LC25000 & General FE \\ %xai

Kashif \etal \cite{kashif2016handcrafted} & 2016 & Colon & Pathology & DET & CRCNuclei & General FE\\

Manivannan \etal \cite{manivannan2017structure} & 2017 & Colon & Pathology & SEG & GlaS & General FE\\

Saba \etal \cite{saba2020brain} & 2020 & Brain & MRI & CLS & BraTS 2015-2017 & General FE \\

Mecheter \etal \cite{mecheter2022deep} & 2022 & Brain & MRI & SEG & Private & General FE \\

TP-CNN \cite{khan2020cascading} & 2020 & Brain & MRI & SEG & BraTS 2014 & General FE \\

SS-OLHF \cite{li2019predicting} & 2019 & Chest & CT & CLS & LIDC-IDRI & General FE \\

Hu \etal \cite{hu2024ct} & 2024 & Chest & CT & CLS & Private & General FE \\ %xai

Prince \etal \cite{prince2025interpretable} & 2025 & Chest & X-ray & CLS & ChestXray8, etc & General FE \\

RC Net \cite{li2019integrating} & 2019 & Eye & OCT & CLS & ZhangLabData & General FE \\

Daoud \etal \cite{daoud2020breast} & 2020 & Breast & US & CLS & BUSData & General FE \\

Sajid \etal \cite{sajid2023breast} & 2023 & Breast & Mammogram & CLS  & CBIS-DDSM & General FE \\

CDHVF \cite{zhang2017classification} & 2017 & Multiple & Multiple & CLS &  ImageCLEF 2016 & General FE \\

% Rajinikanth \etal \cite{rajinikanth2020customized} & 2020 & Brain & MRI & CLS & \cite{menze2014multimodal,clark2013cancer} & General FE \\

% Shankar \etal \cite{shankar2021novel} & 2021 & Chest & X-ray & CLS & \cite{cohen2020covid} & General FE \\

% Huang \etal \cite{huang2022fusing} & 2022 & Prostate & Pathology & CLS & Private & General FE \\

% Joseph \etal \cite{joseph2022improved} & 2022 & Breast & Pathology & CLS & \cite{spanhol2016breast} & General FE \\

% Liu \etal \cite{liu2019fusion} & 2019 & Multiple & Pathology & CLS & \cite{kumar2017comparative, shamir2008iicbu} & General FE \\

% Hasan \etal \cite{hasan2019combining} & 2019 & Brain & MRI & CLS & Private & General FE \\

\midrule

Lou \etal \cite{lou2019image} & 2019 & Chest & CT & CLS & Private & Clinical FE \\ % no xai lacet dh

Chen \etal \cite{chen2025interpretable} & 2025 & Liver & CT & REG & Private & Clinical FE\\

Lassau \etal \cite{lassau2021integrating} & 2021 & Chest & CT & CLS & Private & Clinical FE\\

% Diao \etal \cite{diao2021human} & 2021 & Skin, Breast, Lung, Stomach & Pathology & CLS & \cite{TCGA} & Clinical FE\\
Diao \etal \cite{diao2021human} & 2021 & Multiple & Pathology & CLS & TCGA & Clinical FE\\

% Kapse \etal \cite{kapse2024si} & 2024 & Breast, Lung, Colon & Pathology & CLS & \cite{TCGA} & Clinical FE\\
SI-MIL \cite{kapse2024si} & 2024 & Multiple & Pathology & CLS & TCGA & Clinical FE\\

OvcaFinder \cite{xiang2024development} & 2024 & Ovary & US & CLS & Private & Clinical FE\\

\midrule

Harangi \etal \cite{harangi2019automatic} & 2019 & Eye & Retinal images & CLS & IDRiD & Hybrid FE \\ %no xai

Hagerty \etal \cite{hagerty2019deep} & 2019 & Skin & Dermatology & CLS & HAM10000 & Hybrid FE \\ %no xai

Cruz \etal \cite{cruz2023benign} & 2023 & Breast & Mammogram, US & CLS & mini-DDSM, BUSI & Hybrid FE\\

\bottomrule
\end{tabular}
}
\label{tab:feature}
% \vspace{-1.0em}
\end{table*}

\subsection{Knowledge Graph} \label{subsec:input-2}

A knowledge graph (KG) is a structured representation of factual knowledge that captures relationships between entities in the real world \cite{hogan2021knowledge}. Recent years have witnessed growing interest in integrating structured domain knowledge into downstream applications \cite{ji2021survey}, motivated by the potential of such integration to enhance both the performance and self-explainability of deep learning models. KG-based methods are regarded as S-XAI when the graph is explicitly integrated into the predictive process and helps expose interpretable decision evidence, rather than merely serving as an external information source.
Regarding medical image analysis, knowledge graphs can be broadly classified into three categories based on their construction and knowledge source: 1) Prior KG, which aggregates established domain expertise and clinical knowledge as a foundational resource;
2) Data KG, which is derived from the analysis of large-scale medical image datasets; 
and 3) Hybrid KG, which integrates both prior knowledge and data-driven insights, as illustrated in Fig. \ref{fig:input_exp}b. The list of input explainability methods based on knowledge graph for medical image analysis is presented in Table \ref{tab:kg}.

\subsubsection{Prior knowledge graph}
In the medical domain, prior KGs organize factual information about medical concepts and relationships, constructed from sources such as medical literature, clinical guidelines, and expert opinions. 
By integrating such structured knowledge, AI models can incorporate domain-specific information, enhancing self-explainability and performance across clinical tasks \cite{xie2021survey,liu2024knowledge,liu2023shape}.
In medical VQA, the SLAKE dataset \cite{liu2021slake} provides a prior KG of 52.6k triplets, constructed from OwnThink\footnote{https://www.ownthink.com}. 
Utilizing the SLAKE KG, Guo \etal \cite{guo2022medical} propose identifying the most relevant head entity for a given image-question pair, whereas K-PathVQA \cite{naseem2023k} learns a joint representation that fuses the KG with image-question features for pathology VQA tasks.
Moreover, Huang \etal \cite{huang2023medical} introduce a patient-oriented VQA (P-VQA) dataset, accompanied by a medical KG of 419 entities.
In medical report generation, researchers commonly construct structured knowledge graphs to enhance model performance. Zhang \etal \cite{zhang2020radiology} first build a graph containing the most frequent abnormalities or findings in chest X-rays. Based on this, PPKED \cite{liu2021exploring} utilizes this graph to explore relevant prior knowledge for input X-ray images. Furthermore, Huang \etal \cite{huang2023kiut} and Li \etal \cite{li2024kargen} construct similar symptom graphs that capture correlations among different diseases, thereby generating more reliable and accurate medical reports. Recent approaches, such as KAD \cite{zhang2023knowledge}, integrate prior KGs into the pre-training of medical foundation models, establishing relationships among radiological concepts to enhance generalization to unseen classes.
However, the creation of these KGs largely depends on specialized expertise, which makes the process labor-intensive and time-consuming. Besides, they often lack the adaptability required for analyzing dynamic clinical datasets.

\subsubsection{Data knowledge graph}

Unlike prior KGs, which rely on expert insights, data KGs are built directly from datasets, offering the potential to discover novel relationships and correlations \cite{liu2020hybrid}.
A primary method involves modeling the co-occurrence relationships among entities within a dataset, such as CheXGCN \cite{chen2020label} for multi-label chest X-ray classification and KGAE \cite{liu2021auto} for medical report generation. Intra-data KGs construct graph structures within a single sample, such as representing a whole-slide image as a graph of patches \cite{li2024dynamic} or capturing vascular relations in OCTA images \cite{li2025fine}. In contrast, inter-data KGs capture relationships across different samples. For example, Zheng \etal \cite{zheng2021pay} construct a multimodal KG for COVID-19 diagnosis by connecting various clinical data types \cite{zheng2021pay}.
Further advancing this paradigm, unified frameworks like AG-RCNN \cite{liu2021act}, CCG-Net \cite{zhao2021cross}, and GREN \cite{qi2022gren} integrate both intra-data and inter-data KGs for lesion detection. This dual integration allows models to leverage fine-grained contextual cues within an image alongside population-level relational patterns across cases, thereby enhancing detection accuracy and clinical reasoning.
In summary, constructing data KGs leverages a dataset's inherent characteristics to assist S-XAI models. However, the biases inherent in the source data are also embedded into these KGs, which can vary significantly across datasets and must be critically addressed.

\subsubsection{Hybrid knowledge graph}

Hybrid KGs integrate both prior and data KGs, providing a foundation of established medical facts while dynamically incorporating new information and refining existing knowledge. 
For instance, Zhou \etal \cite{zhou2021contrast} combine intra-data and inter-data KGs to learn abnormal X-ray features, and further leverage a chest radiology prior KG to enhance disease reasoning. 
In medical VQA, Hu \etal \cite{hu2023expert, hu2024interpretable} model spatial and semantic relationships based on prior knowledge while learning implicit relationships directly from data. 
In medical report generation, 
KGVL-BART \cite{kale2023kgvl} first constructs a prior chestX-KG and uses a KG grounding module to extract a subgraph from chestX-KG for each report in the dataset, effectively bridges general medical knowledge with patient-specific contexts.
Moreover, KERP \cite{li2019knowledge}, DCL \cite{li2023dynamic}, and CT-GRAPH \cite{kalisch2025ct} combine both prior structure and data-driven learning. This involves a predefined structure of medical abnormality nodes and relationships, coupled with an adaptive process that learns node features, identifies new nodes, and optimizes edge weights directly from the images.

\textbf{Discussion:} The utilization of medical KGs for S-XAI poses both challenges and promising opportunities. First, KG-based S-XAI should pay closer attention to fusion quality, since weakly aligned graph information may reduce both predictive reliability and explanatory usefulness. Noise robustness is also important, because incomplete or outdated medical knowledge may propagate errors through the fused reasoning process. Furthermore, scalability remains a practical concern for large-scale medical knowledge, where graph size, relation complexity, and update frequency can substantially increase computational and maintenance burden.
Another challenge lies in the heterogeneity of medical data. The expanding variety of data modalities would complicate the effective integration within KGs. Developing robust AI algorithms to extract meaningful features from medical data and align them with medical KGs remains an ongoing research endeavor.

\vspace{1em}
\noindent \textbf{Discussion on information fusion and explanation quality:} At the input explainability level, fusion mainly appears as the integration of handcrafted, clinical, or biological features with deep image representations, as well as the incorporation of prior, data-driven, or hybrid knowledge graphs. These strategies are particularly beneficial for explanation quality when the fused inputs remain clinically meaningful and individually traceable, since this allows the final explanation to be grounded in explicit evidence such as radiomic patterns, semantic attributes, or structured medical knowledge. In this sense, input-level fusion is most supportive of high-quality explanations when it enriches the model with complementary human-understandable information rather than merely increasing feature dimensionality.

% 有数据集名称，无数据集引用
\begin{table*}[t]
\centering
\caption{Input explainability methods based on knowledge graph (KG). The abbreviations here are CLS: classification, DET: detection, MRG: medical report generation, VQA: visual question answering.}
\resizebox{\columnwidth}{!}{
\begin{tabular}{@{}lcccccc@{}}
\toprule

\textbf{Method} & \textbf{Year} & \textbf{Location} & \textbf{Modality} & \textbf{Task} & \textbf{Dataset} & \textbf{KG Type} \\
\midrule

Guo \etal  \cite{guo2022medical} & 2022 & Multiple & X-ray, CT, MRI & VQA & SLAKE & Prior KG  \\

K-PathVQA \cite{naseem2023k} & 2023 & Multiple & Pathology & VQA & PathVQA & Prior KG  \\

MKBN \cite{huang2023medical} & 2023 & Multiple & X-ray, CT, MRI, US & VQA & P-VQA, SLAKE & Prior KG \\

Zhang \etal  \cite{zhang2020radiology} & 2020 & Chest & X-ray & MRG  & ChestXray8 & Prior KG \\

PPKED \cite{liu2021exploring}& 2021 & Chest & X-ray & MRG & IU X-ray, MIMIC-CXR & Prior KG  \\

KiUT \cite{huang2023kiut} & 2023 & Chest & X-ray & MRG & IU X-ray, MIMIC-CXR & Prior KG  \\

KARGEN \cite{li2024kargen} & 2024 &  Chest & X-ray  & MRG & IU X-ray, MIMIC-CXR & Prior KG \\

% KAD \cite{zhang2023knowledge} & 2023 & Chest & X-ray & CLS & MIMIC-CXR, PadChest, ChestXray8, ChestX-Det10 \cite{johnson2019mimic,bustos2020padchest,wang2017chestx,liu2020chestx}  & Prior KG \\
KAD \cite{zhang2023knowledge} & 2023 & Chest & X-ray & CLS & ChestXray8, etc  & Prior KG \\

\midrule

CheXGCN \cite{chen2020label} & 2020 & Chest & X-ray & CLS & ChestXray8, CheXpert& Data KG \\

% Hou \etal  \cite{hou2021multi} & 2021 & Chest & X-ray & CLS & IU-Xray, MIMIC-CXR \cite{demner2016preparing,johnson2019mimic} & Data KG  \\

KGAE \cite{liu2021auto} & 2021 & Chest & X-ray & MRG & IU X-ray, MIMIC-CXR & Data KG \\

WiKG \cite{li2024dynamic} & 2024 & Multiple & Pathology & CLS  & TCGA & Data KG \\

GFT \cite{li2025fine} & 2025 & Eye & OCTA & CLS & OCTA-500 & Data KG \\
 
Zheng \etal \cite{zheng2021pay} & 2021 & Chest & X-ray, CT, US, Text & CLS & Private & Data KG \\

AG-RCNN \cite{liu2021act} & 2021 & Breast & Mammogram & DET & DDSM & Data KG  \\

CCG-Net \cite{zhao2021cross} & 2021 & Chest& X-ray &  DET & ChestXray8 & Data KG  \\

GREN \cite{qi2022gren} & 2022 & Chest & X-ray & DET & ChestXray8 & Data KG  \\

\midrule

Zhou \etal \cite{zhou2021contrast} & 2021 & Chest & X-ray & CLS & ChestXray8, CheXpert & Hybrid KG  \\

% MedKLIP \cite{wu2023medklip} & 2023 & Chest & X-ray & CLS & MIMIC-CXR & Hybrid KG \\

Hu \etal  \cite{hu2023expert} & 2023 & Chest & X-ray & VQA & MIMIC-Diff-VQA  & Hybrid KG  \\

Hu \etal \cite{hu2024interpretable} & 2024 & Chest & X-ray & VQA & MIMIC-CXR-VQA & Hybrid KG \\

KGVL-BART \cite{kale2023kgvl} & 2023 & Chest & X-ray & MRG & IU X-ray & Hybrid KG  \\

KERP \cite{li2019knowledge} & 2019 & Chest & X-ray & MRG & IU X-ray & Hybrid KG  \\

DCL \cite{li2023dynamic} & 2023 & Chest & X-ray & MRG & IU X-ray, MIMIC-CXR & Hybrid KG  \\

CT-GRAPH \cite{kalisch2025ct} & 2025 & Chest & CT &  MRG & CT-RATE & Hybrid KG\\

\bottomrule
\end{tabular}
}
\label{tab:kg}
\vspace{-1.0em}
\end{table*}

\section{Model Explainability} \label{sec:model}

In this section, we present model explainability by designing interpretable model architectures, such as attention-based learning (Sec. 4.1), concept-based learning (Sec. 4.2), and prototype-based learning (Sec. 4.3). 
Attention-based methods often emphasize where the model focuses by highlighting relevant regions or features, whereas concept-based and prototype-based methods are generally more reasoning-oriented. In particular, concept-based methods are designed to clarify what the model learns through semantically meaningful concepts, while prototype-based methods are designed to show what the model matches through representative examples. Based on the above taxonomy, a model is regarded as self-explainable only when the explanatory mechanism is explicitly embedded into the architecture or training objective as a design goal.

% \vspace{-1em}
\subsection{Attention-based Learning} \label{subsec:model-1}

\begin{figure}[t]
\begin{center}
   \includegraphics[width=\linewidth]{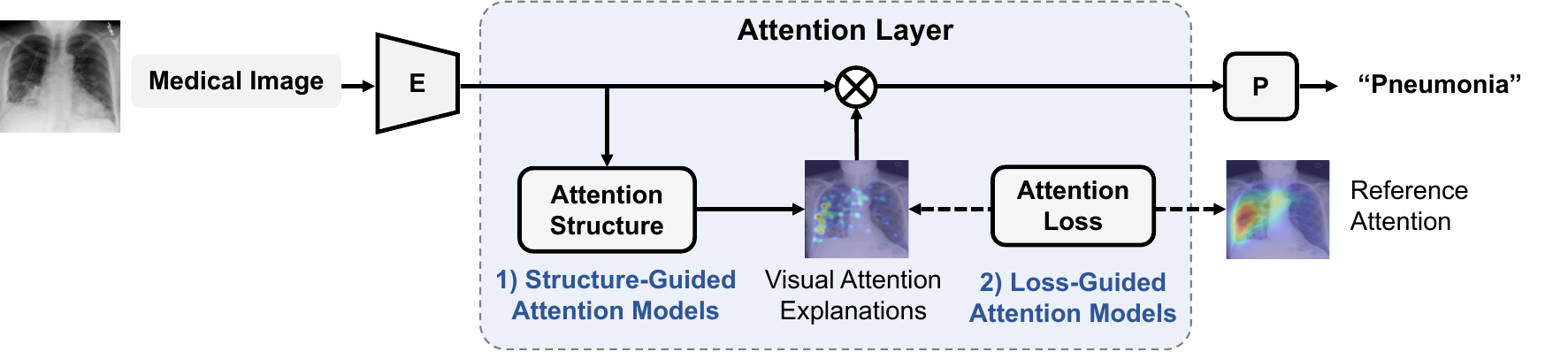}
\end{center}
% \vspace{-1.0em}
   \caption{Attention-based learning, including structure-guided attention models and loss-guided attention models. X-ray images are from RadioTransformer \cite{bhattacharya2022radiotransformer}.}
\label{fig:attention_lsc}
% \vspace{-1.0em}
\end{figure}

Attention-based learning aims to capture specific image regions that are relevant to specific model task while suppressing irrelevant regions based on attention maps \cite{niu2021review}. Therefore, attention-based models can provide visual explanations that interpret model decision-making \cite{salahuddin2022transparency,van2022explainable}. 
We categorize attention-based S-XAI models into structure-guided attention models and loss-guided attention models. As illustrated in Fig. \ref{fig:attention_lsc}, the former estimates attention scores through structural design, whereas the latter constrains attention maps using loss functions.
The list of attention-based learning methods for medical image analysis is shown in Table \ref{tab:attention-based}.

\subsubsection{Structure-guided attention models} 

Structure-guided attention models are characterized by the attention estimator whose outputs directly influence model's predictions. The estimator structure can combine the features from different encoder-layer features to calculate feature attention scores as the generated attention map to effectively explain the model's predictions. For example, 
Jetley \textit{et al.} \cite{jetley2018learn} first introduce attention-based learning for XAI by proposing an attention estimator to calculate feature compatibility scores and weigh feature maps, which are then used directly as input for a linear classifier.

Numerous studies have incorporated multiple attention mechanisms across different sub-fields of medical image analysis.
In medical image classification, Wang \etal \cite{wang2018breast} introduce an attention mechanism for mammography, which reweights feature maps to focus diagnostically relevant mass regions. Furthermore, Li \etal \cite{li2021scouter} propose an explainable slot attention (xSlot) model for fundus images, where an xSlot module iteratively updates attention to produce classification confidence while visualizing positive and negative decision supports. Barata \etal \cite{barata2021explainable} rely on channel and spatial attention methods to guide the network towards relevant features for skin cancer diagnosis.
Moving to medical image segmentation, attention mechanisms are crucial for directing the model's focus toward diagnostically relevant regions, enhancing both the precision of delineation and the interpretability of the results. For instance, Attention Gate \cite{schlemper2019attention} refines segmentation by using coarse features to generate attention maps, enabling finer localization of lesions and improving interpretability in ultrasound and CT scans. CA-Net \cite{gu2020net} integrates spatial, channel, and scale attention, achieving superior segmentation performance on skin lesions and fetal organs while providing clearer visual cues for target area localization. More recently, Anari \etal \cite{anari2025explainable} also integrate multiple attention mechanisms for breast tumor segmentation, effectively enhancing interpretability by precisely highlighting tumor regions.
In brain image analysis, attention-based explainability has been explored for both neurodegenerative \cite{lian2019end} and oncological \cite{tehsin2025gatransformer,hasan2026enhancing} applications.

Beyond 2D data, how to use attention to explain more complex 3D medical image analysis is more challenging. AXIAL \cite{lozupone2024axial} fuses attention weights from sagittal, coronal, and axial slices to diagnose Alzheimer’s disease on 3D MRI brain scans, in which 3D attention maps can be visualized to explain the model's decision-making process.
For fast MRI reconstruction, SDAUT \cite{huang2022swin} proposes a shifted windows deformable attention mechanism which uses reference points to impose spatial constraints on attention and combines the attention outputs from different windows to produce reconstruction results.
Although structure-guided attention models can provide explanations for predictions, they are still difficult to align with clear human-understandable decision-making basis.

\subsubsection{Loss-guided attention models} 

Loss-guided attention models use interpretable labels (i.e., reference attention maps) to constrain generated attention maps from attention mechanism by loss functions. 
Using ground-truth masks of regions of interest (RoIs) to guide the generation of attention maps is a widely adopted approach in medical image classification. 
For pathology images, GuSA \cite{yang2019guided} directly optimizes attention maps using a Dice loss with RoI masks, thereby encouraging the model to focus on relevant regions. To mitigate reliance on costly pixel-level annotations, SAttenReg \cite{yin2021focusing} leverages a pretrained feature extractor to derive clinically informative feature masks, which serve as weak supervision to guide and regularize attention learning. Similar loss-guided attention models have also been extensively explored in dermatology. For instance, Attn-Mel \cite{yan2019melanoma} regularizes attention maps with RoIs, including lesion segmentation or dermoscopic features.  
In medical image segmentation, annotations corresponding to anatomical edges and region shapes are frequently leveraged as auxiliary guidance to refine attention mechanisms for semantic feature extraction. This approach has been demonstrated across diverse imaging modalities, including MRI \cite{sun2020saunet}, dermatology \cite{li2023pmjaf}, and CT \cite{karri2022explainable}, where structural priors help models focus on clinically meaningful regions.
Taking SAUNet \cite{sun2020saunet} as a representative example, the model integrates spatial attention with the attention estimator in U-Net decoders, enabling interpretable feature learning at multiple resolutions. Moreover, SAUNet introduces a gated shape stream operating in parallel to its texture stream. The shape attention maps are explicitly regularized via a binary cross‑entropy loss to align with ground‑truth anatomical boundaries, thereby enhancing both accuracy and edge‑awareness in cardiac MRI segmentation.

Compared with lesion masks, eye tracking data provides a more accurate depiction of expert focus, as it captures the way doctors visually process information during diagnosis.
RadioTransformer \cite{bhattacharya2022radiotransformer} leverages the captured doctors' attention to guide model training. This model employs a teacher-student network to replicate the visual cognitive behavior of doctors when diagnosing diseases on chest radiographs. The teacher model is trained based on the visual search patterns of radiologists, and the student model utilizes an attention loss to predict attention from the teacher network using eye tracking data.

% % 有数据集名称，无数据集引用
\begin{table*}[t]
\centering
\caption{Model explainability methods based on attention-based learning. The abbreviations here are CLS: classification, SEG: segmentation, REC: reconstruction, REG: regression.}
\resizebox{1\textwidth}{!}{
\begin{tabular}{@{}lcccccc@{}}
\toprule
\textbf{Method} & \textbf{Year} & \textbf{Location} & \textbf{Modality} & \textbf{Task} & \textbf{Dataset} & \textbf{Attention} \\

\midrule

Wang \textit{et al.}  \cite{wang2018breast}& 2018 & Breast &  Mammogram & CLS & BCDR-F03 & Structure-Guided\\

SCOUTER  \cite{li2021scouter} & 2021 & Eye &  Retinal images  & CLS &  ACRIMA  & Structure-Guided\\

Barata \textit{et al.}  \cite{barata2021explainable} & 2021 & Skin & Dermatology & CLS &  ISIC 2017-2018 & Structure-Guided\\

% Schempler \textit{et al.}  \cite{schlemper2019attention} & 2019 & Abdominal, Fetal & CT, US &  SEG + DET & \cite{Roth2017Hayashi,roth2016data}  & Structure-Guided\\
Attention Gate  \cite{schlemper2019attention} & 2019 & Multiple & CT, US & SEG, DET & CT-150, Pancreas-CT  & Structure-Guided\\

% Gu \textit{et al.}  \cite{gu2020net} & 2020 & Skin, Fetal & Dermatology, MRI & SEG &  \cite{codella2019skin}  & Structure-Guided\\
CA-Net \cite{gu2020net} & 2020 & Multiple & Dermatology, MRI & SEG &  ISIC 2018 & Structure-Guided\\

Anari \textit{et al.} \cite{anari2025explainable} & 2025 & Breast & US & SEG & BUSI & Structure-Guided\\

Lian \textit{et al.}  \cite{lian2019end} & 2019 & Brain & MRI & REG &  ADNI-1, ADNI-2 & Structure-Guided\\

GATransformer \cite{tehsin2025gatransformer} & 2025 & Brain & MRI & DET & Brain Tumor (Kaggle), etc & Structure-Guided\\

SSPANet \cite{hasan2026enhancing} & 2025 & Brain & MRI & DET & Brain Tumor (Figshare) & Structure-Guided\\

AXIAL \cite{lozupone2024axial} & 2024 & Brain & MRI & CLS & ADNI-1 & Structure-Guided\\

SDAUT \cite{huang2022swin} & 2022 & Brain & MRI & REC & Calgary-Campinas & Structure-Guided\\

\midrule

GuSA \cite{yang2019guided} & 2019 & Breast & Pathology & CLS & BACH & Loss-Guided\\

AttnMel \cite{yan2019melanoma} & 2019 & Skin &  Dermatology & CLS & ISIC 2016-2017 & Loss-Guided\\

SAttenReg \cite{yin2021focusing} & 2021 & Liver &Pathology & CLS & Biopsy4Grading & Loss-Guided\\

SAUNet \cite{sun2020saunet} & 2020 & Cardiac & MRI &  SEG & SUN09, AC17 & Loss-Guided\\

% Karri \textit{et al.}  \cite{karri2022explainable} & 2022 & Skin, Brain, Abdominal & Dermatology, MRI, CT &  SEG & \cite{tschandl2018ham10000,BraTS2020,KAVUR2021101950} & Loss-Guided\\
MSGA-Net \cite{karri2022explainable} & 2022 & Multiple & MRI, CT, Dermatology &  SEG & BraTS 2020, CHAOS, HAM10000 & Loss-Guided\\

PMJAF-Net \cite{li2023pmjaf} & 2023 & Skin &  Dermatology & SEG &  ISIC 2017-2019, HAM10000 & Loss-Guided\\

% RadioTransformer \cite{bhattacharya2022radiotransformer} & 2022 & Chest & X-ray &  CLS &  
% \makecell{RSNA, ZhangLabData, SIIM-FISABIO-RSNA, Public Radiography, COVQU,\cite{shih2019augmenting,kermany2018identifying,lakhani20232021,chowdhury2020can, rahman2021exploring}\\ COVID-19-NY-SBU, MIDRC-RICORD-1C, RICORD, ChestXray8, VinBigData
% \cite{saltz2021stony,tsai2021data, tsai2021rsna,wang2017chestx,nguyen2022vindr}} & Loss-Guided \\
RadioTransformer \cite{bhattacharya2022radiotransformer} & 2022 & Chest & X-ray & CLS & Eye Gaze & Loss-Guided \\
\bottomrule
\end{tabular}
}
\label{tab:attention-based}
% \vspace{-1.0em}
\end{table*}

\textbf{Discussion:} 
Attention-based S-XAI methods provide localization-oriented explanations by highlighting important image regions to guide model predictions.
Structure-guided attention models use attention-weighted outputs as inputs for predictors, reflecting decision-making but often lacking explicit semantics. In contrast, loss-guided attention models offer semantically rich explanations, though their attention maps do not directly affect model decisions. While these approaches help indicate where the model focuses, they are generally less effective at revealing why the prediction is made in a clinically meaningful reasoning sense.

% \vspace{-1em}
\subsection{Concept-based Learning} \label{subsec:model-2}

Concept-based S-XAI methods provide explanations in terms of high-level, human-interpretable attributes rather than low-level, non-interpretable features, enabling users to gain deeper insights into the underlying reasoning with easily understandable concepts \cite{concept_survey}. It also helps in identifying model biases and allows for adjustments to enhance performance and trustworthiness. Most concept-based S-XAI methods focus on making decisions based on a set of concepts while also detailing the contribution of each concept to the final prediction \cite{cbm, pcbm, jain2022extending, tan2024explain}.

\begin{figure}[t]
\begin{center}
   \includegraphics[width=\linewidth]{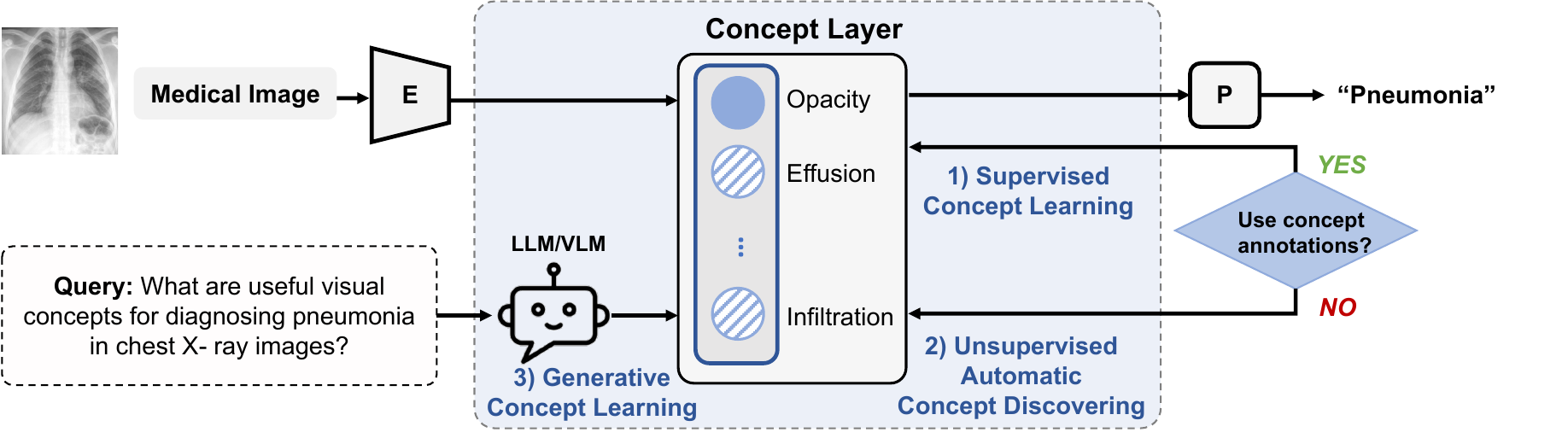}
\end{center}
% \vspace{-1.0em}
   \caption{Concept-based learning, including supervised concept learning, unsupervised automatic concept discovering, and generative concept learning.}
\label{fig:f}
% \vspace{-1.0em}
\end{figure}

The term \textit{Concept} has been defined in different ways, which commonly represents high-level attributes \cite{concept_survey,achtibat2023attribution}. In this paper, we suggest adopting a simple and clear categorization: \textit{Textual Concepts} and \textit{Visual Concepts}. \textit{Textual Concepts} refer to textual descriptions of attributes associated with the classes. For example, in Fig. \ref{fig:f}, the textual concepts for the class ``pneumonia" include terms like \textit{Opacity, Effusion, Infiltration}, etc. \textit{Visual Concepts}, on the other hand, consist of semantically meaningful features within the image that may not be explicitly described in natural language. 
We propose to categorize concept-based S-XAI methods into three types, including supervised concept learning, unsupervised automatic concept discovering, and generative concept learning, as shown in Fig. \ref{fig:f}.
The list of concept-based learning methods for medical image analysis is presented in Table \ref{tab:concept}.

\subsubsection{Supervised concept learning} 
Supervised concept learning methods train deep learning models using annotations of textual concepts, particularly by supervising an intermediate layer to represent these concepts. A notable example is Concept Bottleneck Model (CBM) \cite{cbm}, which first maps latent image features to a concept bottleneck layer, where the number of neurons corresponds to the number of human-defined concepts, and then predicts final results based on the concept scores from this layer. 
Learning concept representations supervised by concept labels enables CBMs to directly show each concept's contribution to the final prediction using the neuron values of the last layer.
Additionally, CBMs allow model editing. When domain experts find certain predicted concept importance unreasonable, they can adjust the model's predictions by modifying the weights of the concept bottleneck layer (i.e., test-time intervention). 
The CBM architecture has inspired many researchers to develop self-explainable methods \cite{cem,barbiero2022entropy,wang2024concept}.

The field of medical image analysis has also advanced a series of CBM-inspired architectures. Interactive CBMs \cite{chauhan2023interactive} can request labels for certain concepts from a human collaborator, which has shown promising interpretability on chest and knee X-ray datasets. The domain of dermatology, benefiting from well-defined diagnostic criteria and publicly available concept-annotated datasets, has emerged as an active area for exploring and validating these concept-based approaches, where CBMs are enhanced by coherent concept-based explanations \cite{patricio2023coherent}, multi-level image-concept alignment \cite{mica}, clinical knowledge integration \cite{pang2024integrating}, concept-driven logical rules \cite{gao2025learning_miccai}, and vision-language model finetuning \cite{zhu2025enhancing_jbhi}.
In ultrasound image classification, MVCBM \cite{marcinkevivcs2024interpretable} extends CBMs to multi-view predictions with incomplete concept sets, whereas SgmaFuse \cite{dai2025interpretable_breast} proposes an multimodal framework that embeds histopathological concepts as intermediate semantic representations.
In contrast to the CBM architecture, some approaches \cite{jalaboi2023dermx,wen2024concept} adopt a multi-task learning paradigm to simultaneously predict both diseases and related concepts. 
Besides, concept whitening based methods \cite{whitening,hou2024concept} aim to whiten the latent space of neural networks, aligning its axes with predefined semantic concepts to improve interpretability. Hybrid neuro-probabilistic reasoning \cite{zhao2021diagnose} integrate clinical concepts with Bayesian networks for verifiable diagnosis.
Nevertheless, supervised concept learning heavily relies on costly expert annotations. This limitation motivates the exploration of unsupervised concept discovery, which learns visual concepts directly from data without manual labels.

\subsubsection{Unsupervised automatic concept discovering}

These models learn to identify visual concepts directly from image features without explicit annotations. The discovered concepts can be visualized and detailed their contributions to final predictions. 
Self-Explaining Neural Networks (SENN) \cite{senn} first utilize a concept encoder to extract clusters of image representations corresponding to different visual concepts, and adopt a relevance parametrizer to calculate the relevance scores of concepts. The final prediction is determined by the combination of discovered concepts and the corresponding relevance scores. 
Following this idea, Sarkar \textit{et al.} \cite{antehoc} utilize both a concept encoder and a concept decoder, which map images into concept space and use the concepts to reconstruct the original images, respectively.

In medical imaging, where expert annotations are costly, unsupervised concept discovery offers a practical path toward explainability. For instance, Visual Concept Mining (VCM) \cite{fang2020concept} addresses the classification of infectious keratitis by first generating visual concepts via saliency-based mining, then integrating both image-level and concept-level features for diagnosis.
While unsupervised concept discovery eliminates the need for manual annotations, the resulting visual concepts are often abstract and not readily translatable into human-understandable language. To combine the interpretability of textual concepts with the efficiency of unsupervised learning, the field has recently shifted toward generative concept learning.

\subsubsection{Generative concept learning} 

This approach leverages foundation models, such as Large Language Models (LLMs) and Vision-Language Models (VLMs), to automatically generate, refine, or align textual concepts with visual content, thereby offering a scalable and semantically rich alternative.
A notable generative concept learning method, namely Language Guided Bottlenecks (LaBo) \cite{labo}, employs GPT-3 \cite{gpt3} to generate textual concepts for each image category, which are filtered to form the concept bottleneck layer. LaBo then uses a pre-trained CLIP \cite{clip} to calculate the similarity between input images and the generated concepts to obtain concept scores. The final prediction is based on the multiplication of a weight matrix and these concept scores.  
Label-free CBM \cite{oikarinen2023labelfree} employs a similar pipeline, but trains an independent network that includes a concept bottleneck layer.  
In the medical domain, Kim \textit{et al.} \cite{kim2023concept} enhance LaBo by incorporating a more fine-grained concept filtering mechanism and conduct explainability analysis. Both leveraging CLIP's vision-language alignment for automated concept scoring in dermatology, MONET \cite{kim2024transparent} trains a large-scale foundation model on medical image–text pairs, whereas Patricio \etal \cite{patricio2024towards} fine-tune CLIP on limited concept annotations. 
Another strategy employs LLMs to generate or retrieve diagnostic-related concepts. 
Liu \etal \cite{liu2023chatgpt} query ChatGPT to produce descriptive symptoms, enriching CLIP's embeddings for zero-shot explainable diagnosis. 
Explicd \cite{gao2024aligning} uses LLMs to define diagnostic criteria across various concept axes. 
Knowledge-enhanced Bottlenecks (KnoBo) \cite{yang2024textbook_knobo} retrieve and formalize medical concepts from textbooks. 
To better align visual features with concepts, MVP-CBM \cite{ijcai2025mvp_cbm} introduces concept preference modeling, capturing how different concepts correlate with features across network layers, and fuses multi-layer activations sparsely to enhance both accuracy and explanation detail.

% 有数据集名称，无引用
\begin{table*}[t]
\centering
\caption{Model explainability methods based on concept-based learning. The abbreviations here are CLS: classification.}
\resizebox{1\textwidth}{!}{
\begin{tabular}{@{}lcccccc@{}}
\toprule
\textbf{Method} & \textbf{Year} & \textbf{Location} & \textbf{Modality} & \textbf{Task} & \textbf{Dataset} & \textbf{Concept} \\
\midrule

CBM  \cite{cbm} & 2020 & Knee & X-ray & CLS & OAI & Supervised (CBM)\\

% Interactive CBM \cite{chauhan2023interactive} & 2023 & Knee, Chest & X-ray & CLS & OAI, CheXpert & Supervised (CBM)\\
Interactive CBM \cite{chauhan2023interactive} & 2023 & Multiple & X-ray & CLS & OAI, CheXpert & Supervised (CBM)\\

Coherent CBE \cite{patricio2023coherent} & 2023 & Skin &  Dermatology & CLS & Derm7pt, PH$^2$ & Supervised (CBM)\\

% \jl{cav} Yan \textit{et al.}   \cite{yan2023towards} & 2023 & Skin &  Dermatology & CLS & Private & Supervised (CBM)\\

MICA \cite{mica} & 2024 & Skin & Dermatology & CLS & Derm7pt, PH$^2$, SkinCon
& Supervised (CBM)\\

% Pang \textit{et al.}  \cite{pang2024integrating} & 2024 & Skin, Blood cell &  Dermatology, Microscopy & CLS & \cite{daneshjou2022skincon,tsutsui2024wbcatt} & Supervised (CBM)\\
Pang \textit{et al.}  \cite{pang2024integrating} & 2024 & Multiple &  Dermatology, Microscopy & CLS & Fitzpatrick 17k, DDI, WBCAtt & Supervised (CBM)\\

% Gao \textit{et al.} \cite{gao2025learning_miccai} & 2025 & Skin, Blood cell & Dermatology, Microscopy & CLS & \cite{daneshjou2022skincon,tsutsui2024wbcatt} & Supervised \\
CRL \cite{gao2025learning_miccai} & 2025 & Multiple & Dermatology, Microscopy & CLS & SkinCon, WBCAtt  & Supervised (CBM)\\

CptAFT \cite{zhu2025enhancing_jbhi} & 2025 & Skin & Dermatology & CLS & Derm7pt, PH$^2$, SkinCon & Supervised (CBM) \\

MVCBM  \cite{marcinkevivcs2024interpretable} & 2024 & Appendix  & US & CLS &  Private  & Supervised (CBM) \\

SgmaFuse \cite{dai2025interpretable_breast} & 2025 & Breast & US & CLS & BUS-BRA, BUSI, BrEaST & Supervised (CBM) \\

% \jl{posthoc}ExAID \cite{lucieri2022exaid} & 2022 & Skin & Dermatology & CLS  & Derm7pt, PH$^2$, ISIC 2019 & Supervised \\

DermX \cite{jalaboi2023dermx} & 2023 & Skin & Dermatology & CLS & DermXDB & Supervised \\

CLAT \cite{wen2024concept} & 2024 & Eye &  Retinal images & CLS & FGADR, DDR & Supervised \\

CAW  \cite{hou2024concept} & 2024 & Skin & Dermatology & CLS & Derm7pt, SkinCon
& Supervised \\

Zhao \textit{et al.}  \cite{zhao2021diagnose} & 2021 &Chest & CT & CLS & LIDC-IDRI & Supervised \\ % not concept discovery

% RC-TPL \cite{zhao2024report_acmmm} & 2024 & Chest & X-ray & CLS & ChestXray8, MIMIC-CXR, CheXpert, etc, RICORD & Supervised \\
% \jl{?}RC-TPL \cite{zhao2024report_acmmm} & 2024 & Chest & X-ray & CLS & ChestXray8, MIMIC-CXR, etc & Supervised \\

\midrule

VCM \cite{fang2020concept} & 2020 & Eye &  Slit lamp microscopy & CLS & Infectious Keratitis  & Unsupervised \\

\midrule

Kim \textit{et al.}  \cite{kim2023concept} & 2023 & Skin & Dermatology & CLS & HAM10000  & Generative \\

% MONET \cite{kim2024transparent} & 2024 & Skin & Dermatology & CLS & ISIC 2019, Fitzpatrick 17k, DDI, SkinCon, Derm7pt & Supervised (CBM)\\
MONET \cite{kim2024transparent} & 2024 & Skin & Dermatology & CLS & ISIC 2019, Fitzpatrick 17k, etc
& Generative \\

Patricio \textit{et al.} \cite{patricio2024towards} & 2024 & Skin & Dermatology & CLS & Derm7pt, PH$^2$, ISIC 2018 & Generative\\

% Xcoop  \cite{xcoop} & 2024 & Multiple & X-ray, Dermatology 
% & CLS & Derm7pt, SkinCon, OpenI, ZhangLabData 
% & Generated concept \\
% XCoOp  \cite{xcoop} & 2024 & Multiple & X-ray, Dermatology 
% & CLS & Derm7pt, SkinCon, OpenI, etc
% & Generative \\

% Liu \textit{et al.}  \cite{liu2023chatgpt} & 2023 & Multiple & Multiple & CLS & ZhangLabData, Montgomery, IDRiD
% & Generative \\
Liu \textit{et al.}  \cite{liu2023chatgpt} & 2023 & Multiple & Multiple & CLS & ZhangLabData, IDRiD, etc
& Generative \\

% Explicd \cite{gao2024aligning} & 2024 & Multiple & Multiple 
% & CLS & HAM10000, NCT, IDRiD, BUSI, MIMIC-CXR
% & Supervised \\
Explicd \cite{gao2024aligning} & 2024 & Multiple & Multiple 
& CLS & HAM10000, NCT, IDRiD, etc
& Generative \\

% KnoBo \cite{yang2024textbook_knobo} & 2024 & Chest, Skin & X-ray, Dermatology & CLS & ChestXray8, CheXpert, MIMIC-CXR, VinDR, HAM10000 & Generated concept  \\
KnoBo \cite{yang2024textbook_knobo} & 2024 & Multiple & X-ray, Dermatology & CLS & ChestXray8, HAM10000, etc & Generative \\

% Wang \textit{et al.} \cite{ijcai2025mvp_cbm} & 2025 & Multiple & X-ray, Dermatology, Histology, Retinal images, US & CLS & ISIC 2018, NCT, IDRiD, BUSI, CMMD, MIMIC-CXR\cite{codella2019skin,nct,porwal2018indian,al20120dataset,cui2021chinese,johnson2019mimic} & Generated concept
% \\ 
MVP-CBM \cite{ijcai2025mvp_cbm} & 2025 & Multiple & Multiple & CLS & ISIC 2018, NCT, IDRiD, etc  & Generative
\\ 

\bottomrule
\end{tabular}
}
\label{tab:concept}
% \vspace{-1.0em}
\end{table*}

\textbf{Discussion:}
Concept-based learning holds significant importance in medical research and applications, particularly in advancing evidence-based medicine. 
However, the lack of fine-grained concept annotations and the performance-explainability trade-off are the limitations of concept-based methods. 
As there are more and more medical foundation models being developed, incorporating knowledge from these models and medical experts to efficiently annotate concept labels for datasets will be a promising and meaningful direction.
However, although general-purpose foundation models have shown satisfactory performance in image-wise concept labeling \cite{labo,oikarinen2023labelfree}, the reliability of medical concept annotation across various medical image modalities remains a significant issue. In particular, LLM- or VLM-based concept generation may introduce hallucinated, weakly grounded, or clinically inappropriate concepts, especially when generated knowledge is insufficiently aligned with image evidence or medical context \cite{asgari2025framework}.
Beyond classification, other medical applications of concept-based approaches should be further explored.

% \vspace{-1em}
\subsection{Prototype-based Learning} \label{subsec:model-3}

\begin{figure}[t]
\begin{center}
   \includegraphics[width=\linewidth]{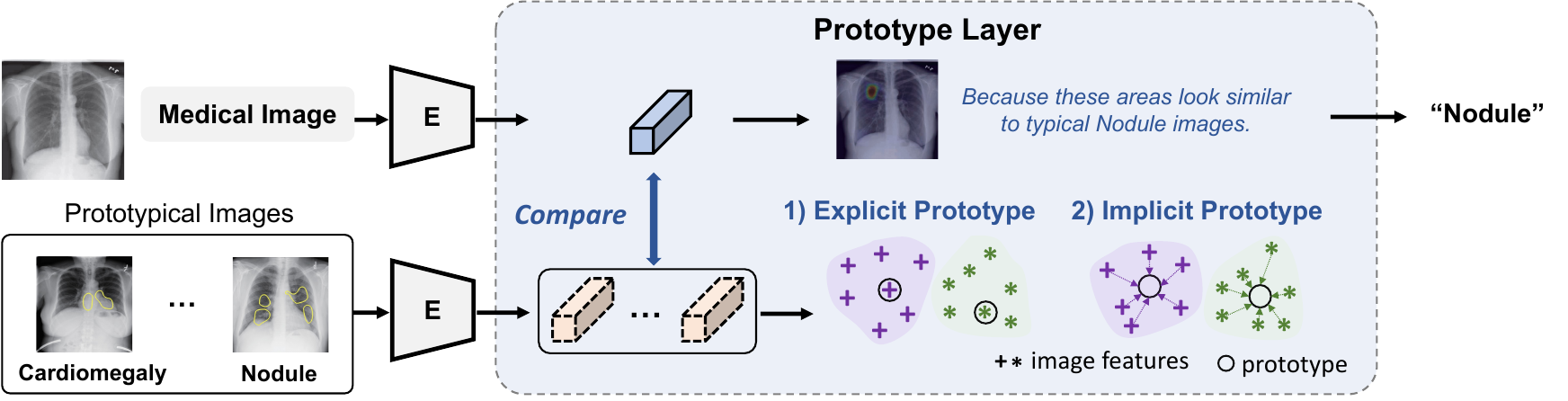}
\end{center}
% \vspace{-1.0em}
   \caption{Prototype-based learning, including explicit prototype and implicit prototype. X-ray images are from XProtoNet \cite{kim2021xprotonet}.}
\label{fig:prototype}
% \vspace{-1.0em}
\end{figure}

Prototype-based S-XAI models aim to provide a decision-making process where a model reasons through comparisons with a set of interpretable example prototypes. This reasoning aligns with human recognition patterns, as humans often identify objects by comparing them to example components \cite{biederman1987recognition}. 
These models first extract features from a given image and then compare the feature maps with the prototypes to calculate similarities. 
This process is considered interpretable because the decision making can be clearly attributed to the contribution of each interpretable prototype (e.g., by the similarity scores).
According to how the prototypes are obtained, we define 
and categorize them to two types, including explicit prototype and implicit prototype, as presented in Fig. \ref{fig:prototype}. Explicit prototypes are specific high-dimensional feature representations extracted from certain training images, whereas implicit prototypes are latent high-dimensional representations that are close to a set of typical images’ representations.
All existing prototype-based S-XAI models do not require supervision at the prototype level and aim to automatically find meaningful prototypes to facilitate interpretable decision making.
Table \ref{tab:prototype} outlines prototype-based learning methods for medical image analysis.

\subsubsection{Explicit prototype based models}
The first model of this type is ProtoPNet \cite{chen2019looks}, which introduces a three-stage training scheme: 
1) Freeze the final layer, and only train the feature extractor; 2) Replace the learned representations in the prototype layer with the nearest feature patch from the training set; 3) Remain the feature extractor fixed and fine-tune the parameters of the final layer.
This scheme has inspired numerous variants that address different limitations of the original framework, such as ProtoShare \cite{rymarczyk2021protopshare}, ProtoPool \cite{rymarczyk2022interpretable}, ProtoDeform \cite{donnelly2022deformable}, ST-ProtoPNet \cite{wang2023learning}, ProtoKNN \cite{ukai2022looks}, and ProtoDebug \cite{bontempelli2023conceptlevel}.

Adopting prototype-based S-XAI models in the medical domain presents additional challenges. In clinical evaluations, while ProtoPNet achieves high interpretability satisfaction among radiologists for mammographic breast‑mass classification \cite{carloni2022applicability}, it has shown performance drops in Alzheimer’s disease detection from MRI \cite{mohammadjafari2021using}.
Unlike natural images where the representative prototype occupies an area with a relatively stable size, medical image features such as disease regions in chest X-ray images can vary significantly in size. 
To address this limitation, XProtoNet \cite{kim2021xprotonet} improves upon ProtoPNet by generating an occurrence map and aggregating similarity scores across the activated region, instead of relying solely on the maximum similarity. Gen-ProtoPNet \cite{singh2021interpretable} introduces prototypes with square and rectangular spatial dimensions for COVID-19 detection in chest X-rays.
In mammography-based breast cancer diagnosis, BRAIxProtoPNet++ \cite{wang2022knowledge} employs knowledge distillation to improve performance, while IAIA-BL \cite{barnett2021case} adopts a multi-stage framework to identify mass margin features, enhancing interpretability and reducing prototype redundancy.
For brain imaging, MProtoNet \cite{pmlr-v227-wei24a} introduces a soft-masking attention module with online CAM loss for 3D multi-parametric MRI to improve interpretability in brain tumor classification. ExPeRT \cite{Hesse_2024_WACV} predicts brain age using a weighted mean of prototype labels. 
ProtoEEG-kNN \cite{tang2025eeg} compares the input image with the prototypes in terms of morphology and spatial distribution for an interpretable classification. X-SiT \cite{bongratz2025x} explores the usage of prototypes in vision transformers in dementia diagnosis. 
In ophthalmology, INSightR-Net \cite{hesse2022insightr} formulates diabetic retinopathy grading as a regression task within a prototype-based framework, and ProtoAL \cite{santos2024protoal} further explores active learning for this setting.
For cardiovascular imaging, ProtoASNet \cite{gu2025protoasnet} learns dynamic spatio‑temporal prototypes that capture shape and motion patterns relevant to aortic stenosis severity. In dermatology, Proto‑RSet \cite{donnelly2025rashomon} generates sets of equivalent prototype networks, allowing clinicians to debug skin‑cancer models without repeated retraining. 
Although explicit prototypes provide clear one‑to‑one mappings between prototypes and image regions, they can obscure which specific visual attribute, such as color, texture, or shape, drives the prototype’s relevance. This limitation has motivated interest in implicit prototype models, which support interpretation through multiple visualizations rather than a single exemplar, offering a more refined understanding of feature contributions.

\subsubsection{Implicit prototype based models}
This type of models follows a similar training scheme as the models based on explicit prototypes, with the major difference in avoiding the prototype replacement step, or only projecting the prototype to the training images' feature patches for visualizations. This scheme is simpler than one that includes prototype replacement step and has different interpretability benefits.
Li \etal \cite{li2018deep} propose the earliest work using latent prototypes, which leverages a decoder to visualize the meanings of the learned prototypes.
EvalProtoPNet \cite{huang2023evaluation} designs a set of loss functions to encourage the learned latent prototypes to be more stable and consistent across different images. 
Recently, to address the concern that prototype-based models often underperform their black box counterparts, Tan \etal \cite{tanpost} develop an automatic prototype discovery and refinement strategy to decompose the parameters of the trained classification head and thus guarantee the performance.

Following the successful demonstration of implicit prototype methods on natural images in works \cite{nauta2021neural,nauta2023pip,ma2024looks,zhu2025interpretable}, recent studies have begun to adapt these techniques to medical image analysis. 
For multi-label classification, CIPL \cite{wang2025cross} leverages implicit prototypes that capture semantics across classes for a more precise and comprehensive understanding of various pathological lesion patterns to disentangle the disease features from multiple classes. In brain tumor diagnosis, ProtoArgNet \cite{ayoobi2025protoargnet} leverages a super prototype, which is a weighted sum of individual prototypes to obtain a unified class prototype for offering both supporting and attacking explanations. 
In OCT classification, PiPViT \cite{oghbaie2026pipvit} introduces a prototypical vision transformer that learns semantically meaningful, patch-based prototypes capable of approximating lesion extent, delivering transparent decision-making and competitive diagnostic performance across multiple retinal OCT datasets.

\textbf{Discussion:}
In terms of performance, implicit prototype based models generally outperform explicit ones, probably due to the greater flexibility in prototype learning. Regarding interpretability, both types of models offer unique advantages. Explicit prototypes can be intuitively explained through one-to-one mappings to the input image, while implicit prototypes can be explained using a diverse set of images with similar activations. 
However, current prototype-based S-XAI models in medical image analysis primarily use explicit prototypes, making the exploration of implicit prototypes in this domain a promising area for future research.

% 保留数据集名字，无引用
\begin{table*}[t]
\centering
\caption{Model explainability methods based on prototype-based learning. The abbreviations here are CLS: classification, REG: regression. }
\resizebox{\textwidth}{!}{
\begin{tabular}{@{}lcccccc@{}}
\toprule
\textbf{Method} & \textbf{Year} & \textbf{Location} & \textbf{Modality} & \textbf{Task} & \textbf{Dataset} & \textbf{Prototype} \\
\midrule

XProtoNet  \cite{kim2021xprotonet} & 2020 & Chest &  X-ray & CLS & ChestXray8 & Explicit\\

Gen-ProtoPNet  \cite{singh2021interpretable} & 2021 & Chest &  X-ray & CLS  & COVID-19 & Explicit \\

Mohammadjafari \etal  \cite{mohammadjafari2021using} & 2021 & Brain & MRI & CLS & OASIS & Explicit \\

Carloni \etal  \cite{carloni2022applicability} & 2022 & Breast & Mammogram & CLS & CBIS-DDSM & Explicit \\
 
BRAIxProtoPNet++  \cite{wang2022knowledge} & 2022 & Breast & Mammogram & CLS & CMMD & Explicit \\

IAIA-BL \cite{barnett2021case} & 2021 & Breast & Mammogram & CLS & Private & Explicit\\
 
MProtoNet \cite{pmlr-v227-wei24a} & 2024 & Brain & MRI & CLS & BraTS & Explicit \\

ExPeRT  \cite{Hesse_2024_WACV} & 2024 & Brain &  MRI, US & REG & IXI, FGLS & Explicit \\

ProtoEEG-kNN  \cite{tang2025eeg} & 2025 & Brain & EEG & CLS & Private & Explicit \\

X-SiT \cite{bongratz2025x} & 2025 & Brain & MRI & CLS & ADNI & Explicit \\ 

INSightR-Net \cite{hesse2022insightr} & 2022 & Eye & Retinal images & REG & EyePACS & Explicit \\

ProtoAL  \cite{santos2024protoal} & 2024 & Eye & Retinal images  & CLS &  Messidor & Explicit \\

ProtoASNet  \cite{gu2025protoasnet} & 2025 & Cardiac & US & CLS & TMED-2 & Explicit\\

% Wang \etal \cite{wang2025cross} & 2025 & Chest and Eye &Thoracic radiography, fundus images&CLS&\cite{wang2017chestx,li2020benchmark} & Implicit \\

Proto-RSet \cite{donnelly2025rashomon} & 2025 & Skin & Dermatology & CLS & HAM10000 &Explicit\\

\midrule
CIPL \cite{wang2025cross} & 2025 & Chest, Eye & Multiple & CLS & ChestXray8, OIA-ODIR & Implicit \\

ProtoArgNet \cite{ayoobi2025protoargnet} & 2025 & Brain & MRI & CLS & Brain Tumor (Kaggle) & Implicit \\

PiPViT \cite{oghbaie2026pipvit} & 2026 & Eye & OCT & CLS & ZhangLabData, OCTID, OCTDL & Implicit \\

% \midrule
% \rowcolor{mygray}\multicolumn{5}{l}{Age prediction}\\
% \midrule

\bottomrule
\end{tabular}
}
\label{tab:prototype}
% \vspace{-1.0em}
\end{table*}

\vspace{1em}
\noindent \textbf{Discussion on information fusion and explanation quality:} At the model explainability level, fusion is reflected in mechanisms such as multi-layer or multi-view attention aggregation, image-concept integration, and prototype-based similarity aggregation. These architectures are more favorable for explanation quality when the fused signals directly contribute to prediction and remain aligned with clinically relevant structures, concepts, or views. In particular, fusion strategies that preserve explicit links between intermediate representations and final decisions are more likely to support faithful and semantically coherent explanations than fusion strategies that simply combine latent features without interpretable structure.

\section{Output Explainability} \label{sec:output}

This section discusses output explainability by generating explanations alongside model predictions, including textual explanations (Sec. \ref{subsec:output-1}) and visual counterfactual explanations (Sec. \ref{subsec:output-2}) .

% \vspace{-1em}
\subsection{Textual Explanation} \label{subsec:output-1}
  
\begin{figure*}[t]
\begin{center}
   \includegraphics[width=\linewidth]{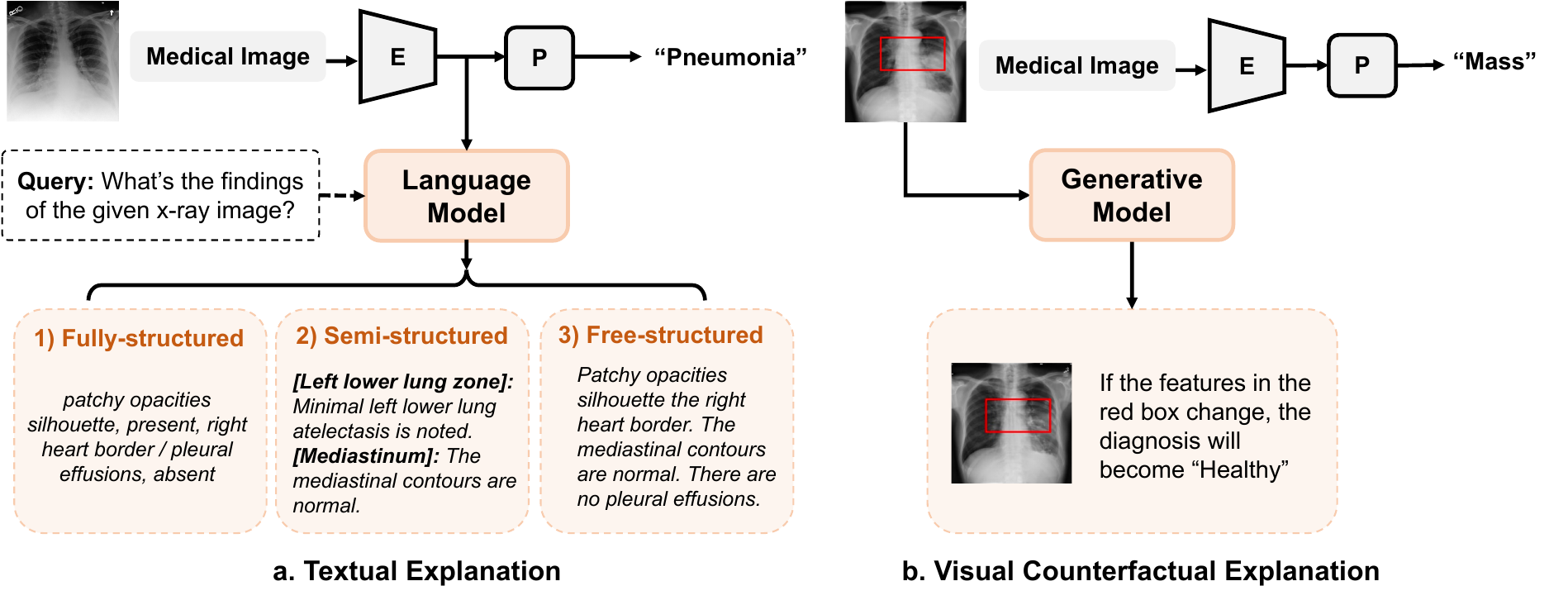}
\end{center}
% \vspace{-1.0em}
   \caption{Output explainability. (a) Textual explanations, provided in fully-structured, semi-structured, and free-structured formats. (b) Visual counterfactual explanations, where the red box highlights the discriminative region between the generated counterfactual and the raw image to provide the visual explanation. X-ray images are sourced from CGN \cite{kim2021interpretation}.
   % between the generated counterfactual image and raw image
   }
\label{fig:output_explanation}
% \vspace{-1.0em}
\end{figure*}

S-XAI models with textual explanation capabilities generate natural language descriptions alongside predictions, similar to image captioning. These methods use text to clarify the model’s reasoning and often demand textual descriptions for training. Textual outputs are regarded as S-XAI when the generated text functions as an explanation of the model’s reasoning or prediction, rather than merely constituting the primary end task itself. Some research further explores the integration of textual and visual explanations. Based on the structure of the generated text, we categorize these approaches into three types, including fully-structured, semi-structured, and free-structured text generation, as illustrated in Fig. \ref{fig:output_explanation}a. Representative methods providing textual explanations in medical image analysis are summarized in Table \ref{tab:text-based}.

\subsubsection{Fully-structured text generation} \label{sec:fully-text}
To address the challenges posed by complex unstructured medical reports, early efforts transformed target texts into fully structured formats, such as descriptive tags \cite{pino2021clinically}, attributes \cite{rodin2019multitask,gale2019producing} with fixed templates, rather than natural language.  
For example, controlled vocabulary terms, such as Medical Subject Headings (MeSH) \cite{lipscomb2000medical}, can be utilized to describe image content instead of unstructured reports. 
Therefore, both Shin \etal \cite{shin2016learning} and Gasimova \etal \cite{gasimova2019automated} employ CNN-RNN architectures to detect diseases and output corresponding MeSH sequences, capturing details such as anatomical location, severity, and affected organs in chest X-rays. 
Nevertheless, natural language descriptions remain more intuitive and comprehensible than isolated tags, motivating subsequent studies to generate reports in a semi-structured textual format.

\subsubsection{Semi-structured text generation} \label{sec:semi-text}
Generating semi-structured text involves a partially structured format with predefined topics and constraints. 
For instance, 
pathology report generation methods \cite{zhang2017mdnet,zhang2017tandemnet,ma2018pathology} produce reports that focus on describing certain types of cell attributes along with a concluding statement.
Additionally, Wang \etal \cite{wang2019computational} introduce a hierarchical framework for medical image explanation, which first predicts semantically related topics and then incorporates these topics as constraints for the language generation model.
More recently, some studies have focused on generating individual sentences based on anatomical regions \cite{wu2chest,li2024anatomical,wang2022inclusive}. For example, Tanida \etal \cite{tanida2023interactive} introduce a Region-Guided Radiology Report Generation (RGRG) method that identifies unique anatomical regions in the chest and generates specific descriptions for the most salient areas, ensuring each sentence in the report is linked to a particular anatomical region. 
Overall, semi-structured approaches effectively balance the rigidity of fully structured reports with the inconsistency of completely free-text reports.

\subsubsection{Free-structured text generation} \label{sec:free-text}

With the advancement of language models, reports generated for a given input image are no longer limited to structured formats; instead, they now focus on more open, free-structured text descriptions. These approaches typically combine an image encoder to extract visual features with a language model to produce coherent sentences \cite{singh2019chest}.
Several research efforts provide comprehensive explanations that include both textual and visual justifications for diagnostic decisions \cite{spinks2019justifying,liu2019clinically,chen2020generating,wang2023metransformer}.

In addition to directly generating textual explanations, some research studies have incorporated the classification of pathological terms or tags in two distinct ways. The first approach utilizes a ``classification-report generation'' pipeline, integrating a classifier within the report generation network to enhance feature representations \cite{yuan2019automatic,lee2019generation}.
In this framework, Yuan \etal \cite{yuan2019automatic} employ a sentence-level attention mechanism alongside a word-level attention model to analyze multi-view chest X-rays, using predicted medical concepts to improve the accuracy of medical reports.
Conversely, the second approach follows a ``report generation-classification'' pipeline, leveraging interpretable region-of-interest characterization for final diagnoses. Illustrating this approach, Zhang \etal \cite{zhang2019pathologist} construct a pathologist-level interpretable diagnostic framework that first detects tumour regions in whole slide images (WSIs), then generates natural language descriptions of microscopic findings with feature-aware visual attention, and finally establishes a diagnostic conclusion. 
Furthermore, integrating region localization and lesion segmentation can enhance the quality of textual explanations \cite{wang2018tienet,jing2018automatic,zeng2020generating,tian2018diagnostic},
as seen in Text-Image Embedding network \cite{wang2018tienet}, which uses multi-level attention to jointly highlight relevant image regions and text words for disease reporting.

The emergence of LLMs and VLMs offers a more interactive and comprehensible method for generating textual explanations. Recent medical VLMs applied to various medical images, such as chest X-rays \cite{thawakar2024xraygpt}, skin images \cite{zhou2024pre}, pathology images \cite{tran2025generating,lu2024multimodal} and general medical images \cite{li2023llava}, can analyze and respond to open-ended questions about the input images, thanks to their pretraining on extensive datasets of image-report pairs.
A prominent example is BiomedGPT \cite{zhang2024generalist}, which utilizes natural-language instructions to perform tasks ranging from disease diagnosis to lesion detection. Moreover, it achieves visual question answering and report generation within an open-ended conversational framework, thereby establishing a versatile paradigm for interpreting medical images.

% 有数据集名，无引用
\begin{table*}[t]
\centering
\caption{Output explainability methods that provide textual explanations. The abbreviations here are MRG: medical report generation, CLS: classification, LOC: location, SEG: segmentation, VQA: visual question answering, Vis: Visual explanation.}
\resizebox{\textwidth}{!}{
\begin{tabular}{@{}lccccccc@{}}
\toprule

\textbf{Method} & \textbf{Year} & \textbf{Location} & \textbf{Modality} & \textbf{Task} & \textbf{Dataset} & \textbf{Text Type} & \textbf{Vis.} \\

\midrule

CNN-TRG \cite{pino2021clinically} & 2021 & Chest & X-ray &  MRG & IU X-ray, MIMIC-CXR  & Fully-structured & $\checkmark$\\

Rodin \etal \cite{rodin2019multitask} & 2019 & Chest &  X-ray &  MRG & MIMIC-CXR &  Fully-structured & $\checkmark$ \\

Gale \etal \cite{gale2019producing} & 2019 & Pelvic &  X-ray &  MRG & Private & Fully-structured & $\checkmark$ \\

Shin \etal \cite{shin2016learning} & 2016 & Chest &  X-ray & CLS, MRG & OpenI &  Fully-structured & -\\

Gasimova \etal \cite{gasimova2019automated} & 2019 & Chest & X-ray & CLS, MRG & OpenI &  Fully-structured & -\\

\midrule

MDNet \cite{zhang2017mdnet} & 2017 & Bladder &  Pathology & MRG & Private  & Semi-structured & $\checkmark$ \\

TandemNet \cite{zhang2017tandemnet}& 2017 & Bladder  & Pathology &  MRG &Private  & Semi-structured & $\checkmark$ \\

Ma \etal \cite{ma2018pathology} & 2018 & Cervix & Pathology &  MRG &Private  & Semi-structured & $\checkmark$\\

Wang \etal \cite{wang2019computational} & 2019 & Chest & X-ray & CLS, MRG &  IU X-ray &  Semi-structured & $\checkmark$ \\

ITA \cite{wang2022inclusive} & 2022 & Chest & X-ray & CLS, MRG & IU X-ray, MIMIC-CXR & Semi-structured & -\\

RGRG \cite{tanida2023interactive} & 2023 & Chest &  X-ray & LOC, MRG & Chest ImaGenome & Semi-structured & $\checkmark$ \\

\midrule

Singh \etal \cite{singh2019chest} & 2019 & Chest & X-ray &  MRG & IU X-ray & Free-structured & - \\

Spinks \etal \cite{spinks2019justifying}  & 2019 & Chest & X-ray &  MRG & IU X-ray & Free-structured & $\checkmark$ \\

Liu \etal \cite{liu2019clinically} & 2019 & Chest & X-ray &  MRG & IU X-ray, MIMIC-CXR & Free-structured & $\checkmark$ \\

Chen \etal \cite{chen2020generating} & 2020 & Chest & X-ray &  MRG & IU X-ray, MIMIC-CXR & Free-structured & $\checkmark$\\

METransformer \cite{wang2023metransformer} & 2023  & Chest & X-ray &   MRG & IU X-ray, MIMIC-CXR & Free-structured & $\checkmark$ \\

Yuan \etal \cite{yuan2019automatic} & 2019 & Chest & X-ray & CLS, MRG &  CheXpert, IU X-ray &  Free-structured & $\checkmark$ \\

Lee \etal \cite{lee2019generation} & 2019 & Breast &  Mammogram & CLS, MRG &  DDSM & Free-structured & $\checkmark$ \\

Zhang \etal \cite{zhang2019pathologist} & 2019  & Bladder & Pathology & CLS, MRG & TCGA & Free-structured & $\checkmark$ \\

TieNet \cite{wang2018tienet} & 2018 & Chest & X-ray & LOC, MRG &  ChestXray8, OpenI &  Free-structured & $\checkmark$ \\

Jing \etal \cite{jing2018automatic} & 2018 & Multiple &  X-ray, Pathology & LOC, MRG & IU X-ray, PEIR Gross &  Free-structured & $\checkmark$ \\ 

SFNet \cite{zeng2020generating} & 2020 & Multiple & US, X-ray & LOC, MRG &  Open-I &  Free-structured & $\checkmark$ \\

Tian \etal \cite{tian2018diagnostic} & 2018 & Abdomen & CT & SEG, MRG & Private & Free-structured & $\checkmark$ \\

XrayGPT \cite{thawakar2024xraygpt} & 2024 & Chest & X-ray & VQA, MRG &  MIMIC-CXR &  Free-structured & - \\

SkinGPT-4 \cite{zhou2024pre} & 2024 & Skin & Dermatology  & VQA, MRG & SkinCon, DermNetNZ &  Free-structured & -\\

HistoGPT \cite{tran2025generating} & 2025 & Skin & Pathology & MRG & HistoGPT-data  & Free-structured & $\checkmark$ \\

PathChat \cite{lu2024multimodal} & 2024 & Multiple & Pathology & VQA, MRG & PMC-OA, TCGA & Free-structured & -\\

LLaVa-Med \cite{li2023llava}& 2023 & Multiple & Multiple  & VQA, MRG & VQA-RAD, SLAKE, PathVQA & Free-structured & - \\

BiomedGPT \cite{zhang2024generalist} & 2024 & Multiple & Multiple & VQA, MRG & IU X-ray, PathVQA, SLAKE & Free-structured & -\\

\bottomrule
\end{tabular}
}
\label{tab:text-based}
\vspace{-1.0em}
\end{table*}

\textbf{Discussion:}
Textual explanations have demonstrated significant effectiveness in providing human-interpretable judgments through natural language. This type of S-XAI approach has become especially valuable with the advancement of language models, enabling the generation of lengthy reports and the ability to answer open-ended questions. However, it is crucial to enhance the quality and reliability of these generated textual explanations. Some recent studies utilize techniques such as knowledge decoupling and instruction tuning to address challenges like hallucination, thereby improving the effectiveness and trustworthiness of textual explanations.

% \vspace{-1em}
\subsection{Visual Counterfactual Explanation} \label{subsec:output-2}

% \begin{figure*}[t]
% \begin{center}
%    \includegraphics[width=\linewidth]{LaTeX/fig/junlin/counterfactual1128.pdf}
% \end{center}
% % \vspace{-1.0em}
%    \caption{Output explainability that provides counterfactual explanations. The difference (red box) indicates the explanation between the generated counterfactual image and raw image. X-ray images are from \cite{kim2021interpretation}. 
%    }
% \label{fig:counterfactual_explanation}
% % \vspace{-1.0em}
% \end{figure*}

Visual counterfactual explanations describe a causal situation by imagining a hypothetical reality that contradicts the observed facts: \emph{If X had not occurred, Y would not have occurred} \cite{christoph2020interpretable}. 
These explanations present a contrastive example: for a given image, its counterfactual image can alter the model's prediction to a predefined output through the minimal perturbation to observations on the original image, as illustrated in the diagram of Fig. \ref{fig:output_explanation}b. Traditional counterfactual explanations are generated through a post-hoc paradigm \cite{schutte2021using}, that is, a classification model is first trained as a black-box model, and then a generative model such as GAN \cite{goodfellow2014generative} is applied to produce the counterfactual counterpart. 
% However, dissociating the model's prediction from its explanation can lead to poor-quality explanations \cite{rudin2019stop}. 
% In particular, 
However, post-hoc counterfactual explanations are susceptible to issues related to the classifier's robustness and complexity (e.g., overfitting and excessive generalization), resulting in explanations that are inadequate for effective interpretability \cite{laugel2019issues}.

To tackle this issue, some research has investigated self-explainable variants of visual counterfactual explanations. An alternative is to incorporate a generative model directly into the predictor, training them jointly so that the model can generate explanations for its own predictions.
% In general, the predictor and explanation generator are trained jointly, hence the presence of the explanation generator is influencing the training of the predictor. 
For example, 
CounterNet \cite{guo2023counternet} combines the training of the predictive model with the generation of counterfactual explanations in an end-to-end framework. Compared to post-hoc approaches, it is able to produce counterfactuals with higher validity.
Similarly, VCNet \cite{guyomard2022vcnet} utilizes a counterfactual generator based on a conditional variational autoencoder, allowing for control and adjustments in the latent space to create more realistic counterfactuals.
In the medical field, Wilms \etal \cite{wilms2021towards} introduce an invertible, self-explainable generative model based on efficient normalizing flow for brain age regression and brain sex classification on 3D neuroimaging data.
This model can generate predictions during the forward process and produce explanations, including voxel-level attribution maps and counterfactual images, in the reverse process.
By integrating data synthesis and predictive tasks into a single model, JointDiffusion \cite{kaleta2025jointdiffusion} enables both components to leverage their shared representations. This synergy directly facilitates the generation of visual counterfactual explanations for various medical images, including malaria, x-rays, and skin data.

\textbf{Discussion:} S-XAI models that generate visual counterfactual explanations alongside predictions show greater promise than post-hoc methods. However, their application in medical image analysis remains largely underexplored, and visual counterfactual explanations have certain limitations. They often rely on model shortcuts rather than producing faithful counterfactuals. Their visual modifications tend to be superficial rather than causally consistent. Most critically, generative models frequently produce anatomically impossible artifacts, undermining clinical plausibility and clinician trust.

\vspace{1em}
\noindent \textbf{Discussion on information fusion and explanation quality:} At the output explainability level, fusion often takes the form of textual-visual explanation integration or joint generative-predictive learning. These strategies can substantially improve explanation quality when complementary explanation channels are semantically coordinated, for example when natural-language rationales, region-level evidence, and counterfactual examples consistently support the same decision. In this setting, information fusion is beneficial not only because it improves user-facing interpretability, but also because it enables richer and more clinically communicative explanations than any single output modality alone.

\section{Evaluation}

Assessing explainability is a crucial component in the development of trustworthy medical AI systems. In this section, we will outline the desirable characteristics and established evaluation methods for explainability.

\subsection{Desirable Characteristics of Explainability} 

It is important for S-XAI models to possess certain desirable qualities when providing explanations. 
In medical applications, the characteristics of high-quality explanations should align closely with the real-world necessities of clinical practice.
Van \etal \cite{van2022explainable} and Adadi \etal \cite{adadi2020explainable} summarize several essential traits of XAI methods for medical image analysis and healthcare, respectively. Jin \etal \cite{jin2023guidelines} propose five criteria for optimizing clinical XAI. 
With the development of LLMs, usability has emerged as a key factor that enhances a model's credibility \cite{wu2024usable}, with interactive and dynamic explanations being preferred over static ones.
Building on the aforementioned literature and the core demands of clinical practice, we propose evaluating clinical S-XAI models across three dimensions, as illustrated in Fig. \ref{fig:evaluation}.

\textbf{Intrinsic trustworthiness} evaluates the fundamental soundness and reliability of the model's internal explanation mechanism. This dimension ensures explanations are not post-hoc rationalizations but are authentically derived from the actual reasoning process, as measured by faithfulness, completeness, and consistency.

\textbf{Extrinsic acceptability} assesses the explanation's practical value for the clinical end-user. It focuses on whether the explanation is useful, credible, and actionable within a clinical workflow, with key metrics including plausibility, understandability, and clinical utility.

\textbf{System feasibility} examines the practical requirements for deploying and sustaining the explainable AI system in real-world clinical environments, through computational efficiency and generalizability.

\begin{figure}[htbp]
\centering
\resizebox{1\textwidth}{!}{
\begin{forest}
for tree={
    grow=0,
    child anchor=west,
    parent anchor=east,
    anchor=west,
    edge path={
        \noexpand\path[\forestoption{edge}]
        (!u.parent anchor) -- +(5pt,0) |- (.child anchor)\forestoption{edge label};
    },
    align=center,
    edge={thick},
    l sep=15pt,
    s sep=10pt,
    if level=0{
        font=\normalsize \bfseries,
        fill=pink!30,
        rounded corners,
    }{
        if level=1{
            font=\normalsize \bfseries,
            fill=cyan!10,
            rounded corners,
        }{
            if level=2{
                font=\normalsize \bfseries,
                fill=orange!10,
                rounded corners,
            }{
                font=\normalsize,
            }
        }
    },
}
[
    Desirable \\ characteristics\\ of S-XAI
    [System \\ feasibility
        [Generalizability
            [Validity when applied to new data or tasks.]
        ]
        [Computational efficiency
            [Time and resources required to generate the explanation.]
        ]
    ] 
    [Extrinsic \\ acceptability
        [Clinical utility
            [Tangible improvement in clinical decision-making or workflow.]
        ]
        [Understandability
            [Ease of comprehension for a non-expert user.]
        ]
        [Plausibility
            [Alignment with human domain knowledge and reasonableness.]
        ]
    ]
    [Intrinsic \\ trustworthiness
        [Consistency
            [Stable and coherent explanations for similar inputs.]
        ]
        [Completeness
            [Coverage of all key factors behind the AI's decision.]
        ]
        [Faithfulness
            [Truthful reflection of the AI's actual decision-making process.]
        ]
    ]
]
\end{forest}
}
\caption{Desirable characteristics of S-XAI.}
\label{fig:evaluation}
\end{figure}

\subsection{Evaluation Methods}
Translating desirable characteristics into measurable metrics requires robust evaluation paradigms. Classic XAI frameworks \cite{doshi2017towards} broadly categorize evaluations into application-grounded, human-grounded, and functionality-grounded methods. However, applying these paradigms in medical contexts demands a more structured alignment with clinical goals. Therefore, we map these evaluation paradigms directly onto the three dimensions of our proposed framework, providing a comprehensive and clinically relevant evaluation guideline.

\subsubsection{Evaluating Intrinsic Trustworthiness}
To evaluate the fundamental soundness and internal reliability of an explanation mechanism, specifically its faithfulness, completeness, and consistency, functionality-grounded evaluations are primarily employed. These evaluations do not involve human-subject investigations; instead, they assess the fidelity of explanations through quantitative metrics. We summarize the specific functionality-grounded metrics utilized to measure intrinsic trustworthiness across different explanation types.

\textbf{Attention-based explanations} assess faithfulness via causal metrics, such as deletion/insertion \cite{petsiuk2018rise} and RemOve And Retrain (ROAR) \cite{hooker2019benchmark}, when ground truth is unavailable.

\textbf{Concept-based explanations} primarily evaluate the faithfulness and completeness of concept representations using metrics such as Concept Error \cite{cbm,antehoc}, T-CAV score \cite{tcav}, Completeness Score \cite{yeh2020completeness}, and Concept Relevance \cite{senn,achtibat2023attribution}. Other metrics including Concept Alignment Score, Mutual Information \cite{cem}, and Concept Purity \cite{wang2023learning} further assess the model's capability to discover coherent and pure concepts.

\textbf{Prototype-based explanations} assess intrinsic prototype quality by two core metrics \cite{nguyen2020quantitative}: non-representativeness (how well selected prototypes capture explanatory patterns) and diversity (the breadth of concepts covered by the prototype set). Beyond these, explanation robustness is quantified via consistency and stability scores, which evaluate the invariance of prototypes across inputs or model runs \cite{huang2023evaluation}.

\textbf{Counterfactual explanations} rigorously evaluate their internal logic and generation quality using mathematical metrics such as counterfactual validity, proximity, sparsity, and diversity \cite{mothilal2020explaining}.

\subsubsection{Evaluating Extrinsic Acceptability}
To evaluate the practical value of an explanation for clinical end-users, specifically its clinical utility, understandability, and plausibility, human-centered evaluations are primarily employed. These evaluations involve domain experts working with real tasks and data \cite{salahuddin2022transparency}. Evaluation approaches include task-based human evaluation, which directly measures human-expert interaction, and proxy-based automated evaluation, which indirectly estimates alignment with expert knowledge.

\textbf{Task-based human evaluation} assesses the practical value for clinical end-users by examining the interaction between the explanation and human domain knowledge \cite{doshi2017towards}. This evaluation ideally takes the form of direct, task-based human studies that recruit medical professionals to interact with explanations. It utilizes qualitative metrics such as the System Causability Scale \cite{holzinger2020measuring} or Likert scales \cite{gale2019producing} to measure subjective understandability, as well as quantitative metrics to track actual clinical utility, including improvements in diagnostic accuracy or reductions in decision time \cite{zhou2021evaluating,sayres2019using}. These clinically oriented metrics are especially important because they reflect whether an S-XAI
system is not only technically explainable, but also practically useful in supporting real clinical decision-making.

\textbf{Proxy-based automated evaluation} circumvents the high cost and time constraints of large-scale human studies by using automated proxies to indirectly estimate plausibility and understandability via alignment with expert ground truth. Visual proxies evaluate attention map plausibility through overlap with expert-drawn semantic masks via Jaccard index or AUC \cite{yan2019melanoma,hou2023diabetic}, radiologist annotations via Activation Precision \cite{barnett2021interpretable}, or clinician eye fixations \cite{muddamsetty2021expert}. Textual proxies apply NLP metrics (BLEU \cite{papineni2002bleu}, ROUGE-L \cite{lin2004rouge}, CIDEr \cite{vedantam2015cider}) to compare generated text with expert clinical reports \cite{patricio2023explainable}. Counterfactual proxies indirectly measure clinical utility via Frechet Inception Distance and Foreign Object Preservation scores \cite{singla2023explaining}.

\subsubsection{Evaluating System Feasibility}
To evaluate the real-world deployment prerequisites of an explanation module, specifically its computational efficiency and generalizability, system benchmarking is primarily employed. In time-sensitive clinical workflows, computational efficiency must be quantified through algorithmic complexity, inference latency, memory footprint, and FLOPs to prevent explanations from imposing prohibitive demands on hospital IT infrastructure \cite{jin2023guidelines}. Furthermore, given the inherent heterogeneity of medical data, multi-center benchmarking is essential for assessing the generalizability of the explanation module, thereby ensuring consistent and reliable explanations under realistic distribution shifts \cite{sun2025explainable}.

\begin{figure}[t]
    \centering
    \includegraphics[width=\linewidth]{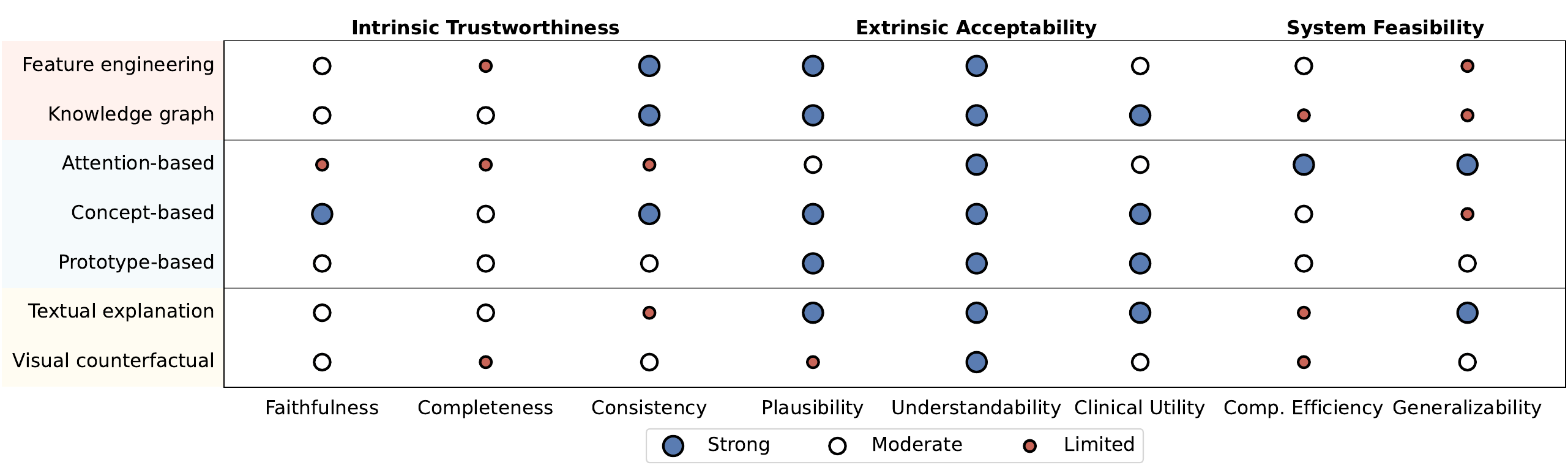}
    \caption{Comparative evaluation of S-XAI methods. The bubble heatmap illustrates the performance of various explanation techniques across three desirable characteristics. Evaluation ratings (limited, moderate, strong) are dual-encoded via bubble size and a divergent red-blue color map.}
    \label{fig:bubble_heatmap}
\end{figure}

\subsection{Comparative Evaluation of S-XAI Methods}

Building upon the desirable characteristics discussed above, we present a comparative evaluation of S-XAI methods, scoring them across eight sub-characteristics, as shown in Fig. \ref{fig:bubble_heatmap}. Each method category was assigned a qualitative rating of limited, moderate, or strong based on the evidence reported in the reviewed papers and the consistency of that evidence across studies: strong, if the reviewed studies consistently report the corresponding evaluation metrics with positive results across multiple datasets or tasks; moderate, if evidence is available but mixed, limited to a single dataset, or only partially addresses the sub-characteristic; limited, if evidence is missing, contradictory, or based solely on anecdotal observations.

\textbf{Intrinsic Trustworthiness:} The internal reliability of S-XAI methods varies significantly across faithfulness, completeness, and consistency. 
Concept-based models exhibit the highest overall trustworthiness by providing explicit reasoning paths and consistent semantic activations, although minor concept leakage limits their completeness. 
Conversely, attention-based models show widespread weaknesses, exhibiting limited faithfulness since attention weights may not reflect true importance, lacking completeness by only highlighting locations, and remaining highly fragile against adversarial noise. 
Input-level methods, including feature engineering and knowledge graphs, offer strong consistency, but their final decision mappings retain black-box elements that compromise faithfulness. 
Regarding generative approaches, modern aligned textual explanations achieve moderate faithfulness by mitigating hallucinations, whereas visual counterfactuals often reveal model shortcuts rather than true causal mechanisms. Furthermore, both prototype and counterfactual methods struggle with completeness, as their reliance on limited cases or minimal modifications obscures other contributory factors.

\textbf{Extrinsic Acceptability:} Most S-XAI methods demonstrate strong understandability by successfully translating opaque model weights into interpretable formats. However, their practical value is primarily differentiated by plausibility and clinical utility. 
Textual explanations and prototype-based models offer exceptional clinical utility by seamlessly integrating into clinical workflows and aligning directly with physicians' traditional case-based reasoning habits. 
Input-level methods and concept-based models also achieve strong plausibility by utilizing standard medical terminology. Within this group, manual feature engineering provides slightly less overall utility compared to end-to-end concept graphs. 
In contrast, attention maps offer only moderate utility restricted to basic lesion localization and occasionally reduce plausibility by highlighting irrelevant backgrounds. Visual counterfactuals face the most severe acceptability challenges, as generative models frequently produce anatomically impossible artifacts that severely degrade medical plausibility and clinician trust.

\textbf{System Feasibility:} Deployment readiness depends primarily on computational efficiency and generalizability to new data or tasks. 
Attention-based models excel in both aspects, functioning as a plug-and-play, lightweight module applicable to diverse medical imaging tasks. Textual explanations leverage the zero-shot capabilities of modern VLMs to generalize across unseen diseases, yet their high GPU requirements and inference latency limit their use in time-sensitive emergency care. 
Conversely, methods reliant on domain-specific manual design, including feature engineering, knowledge graphs, and concept-based models, face severe generalizability bottlenecks. Since their interpretable elements are highly customized, they struggle to maintain validity across different diseases or imaging modalities without extensive re-annotation. 
Finally, prototype and visual counterfactual methods demonstrate moderate feasibility. They are constrained by the computational overhead of distance calculations or generative inference, and their explanation validity remains highly sensitive to cross-institutional data shifts.

\section{Challenges and Future Directions}

Despite the rapid advancements in S-XAI for medical image analysis, several significant challenges remain unresolved. 
In this section, we analyze the existing challenges and discuss future directions from three perspectives, including data and evaluations, methodological advances, and clinical translation. Following this progression, benchmark construction remains a prerequisite for trustworthy comparison, algorithmic progress is currently driven by topics including diagnostic performance-interpretability balance, foundation models, and multimodal or causal reasoning, while human-in-the-loop interaction serves as an important bridge from model design to deployment. Overall, these directions collectively point toward the development of S-XAI systems that are not only technically explainable, but also clinically usable and translatable in real-world medical settings.

\subsection{Benchmark Construction for Clinically Meaningful S-XAI}

Establishing benchmarks for S-XAI in medical image analysis is essential, which enables standardized evaluations and fair comparisons between different methods.

\textbf{Dataset.}
A major challenge of dataset construction is the limited availability of doctors to annotate large-scale medical images, especially in S-XAI, which requires fine-grained annotations such as concepts and textual descriptions. Moreover, inter-observer variability directly affects annotation quality, as clinicians may differ in how they describe concepts, delimit lesions, or summarize findings. Current medical datasets that meet interpretability standards often have limited volumes, reducing the generalizability and applicability of S-XAI methods in real-world contexts.

\textbf{Evaluation.}
Automated evaluation of explanations poses another significant challenge. In the medical field, human-centered evaluations often depend on clinician expertise, but variability in expert opinions can lead to biased and subjective assessments \cite{tonekaboni2019clinicians}. Additionally, existing functionality-grounded evaluations still rely on manual annotations. Furthermore, benchmark performance may vary substantially across hospitals, devices, and imaging protocols, necessitating multi-center and multi-device validation. Most S-XAI studies are still evaluated mainly through technical experiments on public benchmarks, with limited explicit clinical validation.

To address these challenges, future research should explore: (1) semi-automated annotation frameworks that assist clinicians while preserving validity; (2) systematic evaluation protocols with objective interpretability metrics and clinically meaningful benchmark settings; and (3) multi-center benchmark construction with expert-involved, workflow-oriented validation to ensure robustness under realistic clinical heterogeneity.

\subsection{Balancing Diagnostic Performance and Interpretability}

It is commonly believed that as model complexity increases to boost performance, interpretability tends to decline \cite{gunning2019darpa,minh2022explainable}. Conversely, more interpretable models may sacrifice prediction accuracy. In medical image analysis, this trade-off carries high stakes because small performance changes can have clinical consequences, while weak or abstract explanations fail to support clinician trust.

However, some researchers argue that \textit{there is no scientific evidence for a general trade-off between accuracy and interpretability} \cite{rudin2022interpretable}. 
Recent advancements in concept-based models have shown performance comparable to black-box models in medical image analysis, particularly for tasks where clinically meaningful signals can be expressed through concepts, regions, or prototypes (e.g., lesion characterization, pathology pattern recognition, anatomically grounded reporting). Yet this balance may vary across modalities, tasks, and data regimes, especially when disease presentation is subtle, heterogeneous, or weakly annotated.

Future S-XAI methods should therefore move beyond a generic discussion of trade-off and instead investigate when interpretability can enhance robustness, generalizability, and clinical usability without sacrificing predictive quality. Establishing task-specific evidence along this direction will be essential for developing clinically trustworthy S-XAI systems.

\subsection{Foundation Models in Medical Image S-XAI}

Medical foundation models, such as LLMs \cite{singhal2025toward,wu2024pmc,gu2021domain} and VLMs \cite{moor2023foundation,moor2023med}, are trained on extensive medical corpora, allowing them to inherently encode rich domain-specific knowledge. The convergence of S-XAI with these large models creates substantial opportunities for developing trustworthy medical AI systems in the future \cite{wu2024usable}.

\textbf{S-XAI benefits foundation models.}
Foundation models are typically large, with an enormous number of parameters, making it challenging to explore their decision-making processes. This complexity can lead to potential biases and a lack of transparency. In addition to using post-hoc techniques \cite{ye2022can,wu2024language} to interpret foundation models, S-XAI methods can enhance input explainability through explainable prompts \cite{wei2022chain,xcoop}. More broadly, S-XAI enhances the transparency and trustworthiness of foundation models by aligning explanations with internal reasoning, grounding outputs in clinically meaningful concepts. This helps reduce hallucinations and mitigate black-box risks in medical foundation models.

\textbf{Foundation models advance S-XAI.}
Foundation models learn useful representations from the clinical knowledge in medical corpora \cite{singhal2023large}. By leveraging their advanced capabilities, S-XAI methods can generate user-friendly explanations, support open-ended visual question answering, and enable flexible generative concept-based learning \cite{singh2023augmenting}. In addition, multimodal foundation models facilitate concept discovery and reasoning, thereby expanding self-explainability beyond handcrafted concepts or explicitly designed structures. These capabilities are especially promising for applications such as report generation, interactive explanation, and clinical decision support.
Additionally, foundation models can aid in evaluating S-XAI methods that mimic human cognitive processes \cite{bills2023language}.

The use of foundation models in S-XAI also introduces important challenges. Even when they improve semantic richness and user-facing acceptability, the faithfulness of generated explanations to the model’s internal reasoning is difficult to guarantee. Pretrained foundation models may also inherit dataset bias, domain mismatch, and hallucination risks, particularly problematic in high-stakes medical settings. Their large scale raises practical concerns regarding computational cost, reproducibility, and reliable deployment. Thus, foundation models for medical S-XAI should be evaluated not only on expressive capability, but also on trustworthiness, clinical safety, and robustness.

\subsection{Multimodal and Causal Explainability in Medical Imaging}

\textbf{Multimodal explainability.}
Medical data often exists in various forms, such as images, texts, and omics. By integrating these modalities, multimodal S-XAI methods can provide more comprehensive explanations that align with clinicians' processes. Additionally, S-XAI methods can be used to identify correlations during multimodal data fusion \cite{xu2023multimodal}, offering significant potential for discovering new biomarkers and for revealing how different clinical evidence sources jointly support model predictions. Multimodal explanations also help reveal when evidence across modalities is not fully aligned, for example, through modality-specific weighting, graph-based relational reasoning, or inconsistencies between textual and visual explanations. Future research should therefore pay greater attention to conflict-aware multimodal S-XAI that can surface competing evidence across modalities, not only supportive correlations.

\textbf{Causal explainability.}
Another direction for S-XAI focuses on causality, which defines the cause-and-effect relationship and can be mathematically modeled \cite{pearl2009causality}. 
Traditional deep learning methods in medical imaging often confuse correlation with causation, leading to potentially harmful errors when models rely on scanner-dependent artifacts, acquisition shortcuts, or confounded population patterns instead of genuine disease mechanisms.
For instance, DeGrave \etal \cite{degrave2021ai} found that COVID-19 diagnosis methods are primarily identifying spurious correlations using XAI techniques. Several efforts \cite{castro2020causality,luo2022pseudo} have been made to address dataset biases, which would be highly beneficial.
Future S-XAI research should therefore place greater emphasis on causal analysis that can distinguish clinically meaningful factors from superficial correlations, especially in high-stakes tasks such as diagnosis, risk prediction, and treatment assessment.

\subsection{Human-in-the-loop S-XAI for Expert-guided Medical Imaging}

Integrating Human-in-the-loop (HITL) processes is crucial for effectively implementing S-XAI in the medical field, 
as the interpretation of disease findings depends on subtle visual patterns, anatomical context, and expert judgment that cannot be fully captured by static annotations. This approach not only enhances the overall performance of AI systems but also fosters trust among medical experts.

\textbf{Improving accuracy through human intervention.}
A HITL framework allows for the identification and removal of potential confounding factors, such as artifacts or biases in datasets.
For instance, clinicians can correct concept labels \cite{yan2023towards}, resolve region-level reasoning ambiguities, and refine attribution maps. This collaborative approach significantly enhances the prediction accuracy by incorporating expert insights.

\textbf{Enhancing explainability through human feedback.} 
A versioning or feedback evaluation system should be embedded into real diagnostic workflows, enabling clinicians to assess whether explanations are sufficiently localized, semantically meaningful, and actionable for routine reporting or decision support.
This requires fostering collaboration between S-XAI researchers and clinical practitioners to systematically gather and utilize feedback for model refinement \cite{tonekaboni2019clinicians,cai2019human}. The quality of this feedback loop directly determines whether S-XAI explanations are clinically usable and trusted.

A challenge in integrating HITL processes is the variability in clinician expertise and availability, which can impact the consistency and quality of feedback, especially when radiologists, pathologists, or other specialists focus on different visual cues or reporting habits. Ensuring effective integration of human knowledge into the AI training process without introducing biases or errors is a complex task.

\subsection{Clinical Deployment and Real-world Validation of S-XAI}

Beyond methodological advances, the clinical value of S-XAI ultimately depends on whether these systems can be deployed reliably in real medical environments. In practice, explainable models must be compatible with existing diagnostic workflows, provide explanations at an appropriate level of detail for clinicians, and support calibrated trust rather than either blind reliance or systematic rejection \cite{ghassemi2021false}. In addition, deployment in hospitals requires S-XAI systems to remain robust under institution-level domain shifts caused by differences in scanners, imaging protocols, patient populations, and annotation styles.

Future research should therefore prioritize deployment-oriented development \cite{muehlematter2021approval,benjamens2020state}, including multi-center and prospective validation, sustainable clinician feedback loops, efficient updating and monitoring mechanisms, and explicit consideration of computational cost and maintenance burden. Workflow-aware explanation design, while already implied by the need for compatibility, should also be systematically evaluated. Moreover, S-XAI systems must be designed to meet the transparency, traceability, and human oversight requirements for regulatory and ethical accountability in Software as a Medical Device (SaMD) contexts \cite{act2024eu,fda2017samd}. Such regulatory alignment is a precondition for clinical adoption rather than an afterthought. These efforts are essential for translating S-XAI from research prototypes into trustworthy clinical decision-support systems.

\section{Conclusion}

This survey reviews recent advancements in self-explainable artificial intelligence (S-XAI) for medical image analysis. Contrary to previous surveys that primarily focus on post-hoc XAI techniques, this paper emphasizes inherently interpretable S-XAI models, which are gaining traction in research.
This survey introduces S-XAI from three key perspectives, i.e., input explainability, model explainability, and output explainability. 
Additionally, this survey explores the desired characteristics of explainability and various evaluation methods for assessing explanation quality. While significant progress has been made, it also highlights key challenges that need to be tackled and provides insights for future research on trustworthy AI systems in clinical practice.
Finally, this survey highlights the close connection between information fusion and S-XAI. Fusion operates at all three levels by integrating explainable inputs, aggregating interpretable internal representations, and unifying explanatory outputs. Explanation quality depends on how well these sources are fused, and keeping the fused signals explicitly traceable to the final decision leads to more faithful explanations. The fusion perspective thus offers a unifying lens for understanding, designing, and evaluating S-XAI methods.
Overall, this survey serves as a valuable reference for the XAI community, particularly within the medical imaging field, and lays the groundwork for future advancements that will improve the transparency and trustworthiness of AI tools in healthcare.

\section{Acknowledgments}
This work was supported by the Hong Kong Innovation and Technology Fund (Project No. MHP/002/22), HKUST (Project No. FS111) and Research Grants Council of the Hong Kong (No. R6003-22 and T45-401/22-N).

% \appendix
% \input{appendix}

\bibliographystyle{elsarticle-num}
\bibliography{junlin,yequan, hongmei,andong,sicen}

\end{document}